\documentclass{article}

     \PassOptionsToPackage{numbers, compress}{natbib}

 \usepackage{neurips_2019}

\usepackage[utf8]{inputenc} 
\usepackage[T1]{fontenc}    
\usepackage{hyperref}       
\usepackage{url}            
\usepackage{booktabs}       
\usepackage{amsfonts}       
\usepackage{nicefrac}       
\usepackage{microtype}      

\usepackage[table]{xcolor}

\usepackage{amsmath}
\usepackage{amsfonts}
\usepackage{amssymb}
\usepackage{wrapfig}
\usepackage{subcaption}
\usepackage{multirow}
 \usepackage{mathtools} 

\usepackage{verbatim}

\usepackage{anyfontsize}

\usepackage{color}
\usepackage{tikz}
\usetikzlibrary{arrows,shapes,snakes,automata,backgrounds,fit,petri}
\usepackage{adjustbox}

\newcommand{\RN}[1]{%
	\textup{\lowercase\expandafter{\it \romannumeral#1}}%
}

\makeatletter
\newcommand{\distas}[1]{\mathbin{\overset{#1}{\kern\z@\sim}}}%

\usepackage{enumitem}

\usepackage[lined,boxed,commentsnumbered,ruled,linesnumbered]{algorithm2e}

\newcommand{\ie}[0]{\emph{i.e., }}

\newcommand{\eg}[0]{\emph{e.g., }}

\newcommand{\beq}{\vspace{0mm}\begin{equation}}
\newcommand{\eeq}{\vspace{0mm}\end{equation}}
\newcommand{\beqs}{\vspace{0mm}\begin{eqnarray}}
\newcommand{\eeqs}{\vspace{0mm}\end{eqnarray}}
\newcommand{\barr}{\begin{array}}
\newcommand{\earr}{\end{array}}

\newcommand{\Cmat}{{\bf C}}

\newcommand{\Imat}{{\bf I}}

\newcommand{\Vmat}[0]{{{\bf V}}}

\newcommand{\Xmat}[0]{{{\bf X}}}

\newcommand{\bv}[0]{{\boldsymbol{b}}}

\newcommand{\hv}[0]{{\boldsymbol{h}}}

\newcommand{\xv}{\boldsymbol{x}}

\newcommand{\zv}{\boldsymbol{z}}

\newcommand{\thetav}{\boldsymbol{\theta}}

\newcommand{\muv}[0]{{\boldsymbol{\mu}}}

\newcommand{\piv}{\boldsymbol{\pi}}

\newcommand{\sigmav}[0]{{\boldsymbol{\sigma}}}

\newcommand{\phiv}{\boldsymbol{\phi}}
\newcommand{\psiv}{\boldsymbol{\psi}}

\newcommand{\Lcal}{\mathcal{L}}
\newcommand{\Ncal}{\mathcal{N}}

\newcommand{\Bcal}{\mathcal{B}}

\newcommand{\Kcal}{\mathcal{K}}

\newcommand{\Hcal}{\mathcal{H}}

\newcommand{\Ucal}{\mathcal{U}}

\ifx\assumption\undefined
\newtheorem{assumption}{Assumption}
\fi

\ifx\definition\undefined
\newtheorem{definition}{Definition}
\fi

\ifx\remark\undefined
\newtheorem{remark}{Remark}
\fi

\title{RaCT: Towards Amortized Ranking-Critical Training for Collaborative Filtering}

\author{%
}

\begin{document}

\maketitle

\begin{abstract}
Collaborative filtering is widely used in modern recommender systems. Recent research shows that variational autoencoders (VAEs) yield state-of-the-art performance by integrating flexible representations from deep neural networks into latent variable models, mitigating limitations of traditional linear factor models. VAEs are typically trained by maximizing the likelihood (MLE) of users interacting with ground-truth items.
While simple and often effective, MLE-based training does not directly maximize the recommendation-quality metrics one typically cares about, such as top-$N$ ranking. 
In this paper we investigate new methods for training collaborative filtering models based on actor-critic reinforcement learning, to directly optimize the non-differentiable quality metrics of interest. Specifically, we train a {\it critic} network to approximate ranking-based metrics, and then update the {\it actor} network (represented here by a VAE) to directly optimize against the learned metrics.
%
%
In contrast to traditional learning-to-rank methods that require to re-run the optimization procedure for new lists, our critic-based method amortizes the scoring process with a neural network, and can directly provide the (approximate) ranking scores for new lists.
Empirically, we show that the proposed methods outperform several state-of-the-art baselines, including recently-proposed deep
learning approaches, on three large-scale real-world datasets. 
\end{abstract}

\section{Introduction}
\vspace{-3mm}
Recommender systems are an important means of improving a user's web experience.
%
Collaborative filtering is a widely-applied technique in recommender systems~\cite{ricci2015recommender}, in which patterns across similar users and items are leveraged to predict user preferences~\cite{su2009survey}. This naturally fits within the learning paradigm of latent variable models (LVMs)~\cite{bishop2006pattern}, where the latent representations capture the shared patterns. Due to their simplicity and effectiveness, LVMs are still a dominant approach. However, traditional LVMs employ linear mappings of limited modeling capacity~\cite{paterek2007improving,mnih2008probabilistic}, which may yield suboptimal performance, especially for large datasets~\cite{he2017neural}.
This problem has been mitigated recently in a growing body of literature that involves applying deep neural networks (DNNs) to collaborative filtering~\cite{he2017neural,wu2016collaborative,liang2018variational}.
Among them, variational autoencoders (VAEs)~\cite{kingma2013auto,rezende2014stochastic} have been proposed as non-linear extensions of LVMs~\cite{liang2018variational}. Empirically, they significantly outperform state-of-the-art methods. One essential contribution to the improved performance is the use of the multinomial likelihood, which is argued to be a close proxy to the ranking loss.

This is desirable, because we generally care most about the ranking of predictions in recommender systems. 
%
Hence, prediction results are often evaluated using top-$N$ ranking-based metrics, such as Normalized Discounted Cumulative Gain~\cite{jarvelin2002cumulated}. The VAE is trained to maximize the likelihood of observations; as shown below, this does not necessarily result in higher ranking-based scores. 
A natural question concerns whether one may optimize directly against ranking-based metrics (or a close proxy of them). 
Previous work on learning-to-rank has been explored in the information-retrieval community, where relaxations/approximations are considered~\cite{weimer2008cofi,liu2009learning,li2014learning,weston2013learning}. They are not straightforward to adapt for collaborative filtering, especially for large-scale datasets. 
Since these methods generally focus on the pair-wise ranking loss between positive and negative items, it is computationally expensive to have a low-variance approximation to the full loss when the number of items is large.

In this paper, we borrow the actor-critic idea from reinforcement learning (RL)~\cite{sutton1998reinforcement} to propose an efficient and scalable learning-to-rank algorithm. The critic is trained to approximate the ranking metric, while the actor is trained to optimize against this learned metric. 
Specifically, with the goal of making the actor-critic approach practical for recommender systems,
we introduce a novel feature-based critic architecture. Instead of treating raw predictions as the critic input, and hoping the neural network will discover the metric's structure from massive data, we consider engineering sufficient statistics for efficient critic learning.
%
Experimental results on three large-scale real-world datasets demonstrate that the proposed method significantly improves on state-of-the-art baselines, and outperforms other recently proposed neural-network approaches. 


\vspace{-2mm}
\section{Preliminaries: VAEs for Collaborative Filtering}
\vspace{-2mm}
Vectors are denoted as bold lower-case letters $\xv$, matrices as bold uppercase letters $\Xmat$, and scalars as lower-case non-bold letters $x$. We use $\circ$ for function composition, $\odot$ for the element-wise multiplication, and $| \cdot | $ for cardinality of a set. $\delta( \cdot  )$ is the indicator function.

We use $n \in \{1, \dots , N \}$ to index users, and $m \in \{1, \dots, M \}$ to index items. The user-item interaction matrix $\Xmat \in \{0,1\}^{N \times M }$ collected from the users' implicit feedback is defined as:
\begin{align}
x_{nm}
\left\{\begin{matrix}
1, & \text{if interaction of user}~n~\text{with item}~m~\text{is observed};\\ 
0, & \hspace{-54mm}\text{otherwise} 
\end{matrix}\right.
\end{align}
Note that $x_{nm}=0$ does not necessarily mean that user $n$ dislikes item $m$; it can be that the user is not aware of the item. Further, $x_{nm}=1$ is not equivalent to saying user $n$ likes item $m$, but at least there is interest. 

%

\paragraph{VAE model}~\hspace{-4mm}
VAEs have been investigated for collaborative filtering \cite{liang2018variational}, where this principled Bayesian approach is shown to be the state-of-the-art for large-scale datasets.
Given the user's interaction history $\xv = [x_1, . . . , x_M ]^{\top} \in \{0,1\}^M$, our goal is to predict the full interaction behavior with all remaining items. 
To simulate this process during training, a random binary mask $\bv \in \{0,1\}^M$ is introduced, with the entry $1$ as {\it un-masked}, and $0$ as {\it masked}. Thus,  $\xv_h = \xv \odot  \bv$ is the user's partial interaction history. The goal becomes recovering the masked interactions: $\xv_p= \xv \odot  (1-\bv)$.

In LVMs, each user's binary interaction behavior is assumed to be controlled by a $k$-dimensional user-dependent latent representation $\zv \in \R^K$. When applying VAEs to collaborative filtering~\cite{liang2018variational}, the user's latent feature $\zv$ is represented as a distribution $q(\zv | \xv )$, obtained from some partial history $\xv_h$ of $\xv$. With the assumption that $q(\zv | \xv )$ follows a Gaussian form, the {\em inference} of $\zv$ for the corresponding $\xv$ is performed as:
\begin{align} \label{eq_inference} \hspace{-10mm}
 q_{\phiv}(\zv | \xv ) =  \Ncal(\muv, \mbox{diag}(\sigmav^2)),
 ~~\text{with}~~ \muv, \sigmav^2=f_{\phiv}(\xv_{h}),~~
 \xv_h = \xv \odot  \bv,~~\hspace{1mm} \bv \sim \mbox{Ber}(\alpha)
\end{align}
where $\alpha$ is the hyper-parameter of a Bernoulli distribution, $f_{\phiv}$ is a $\phiv$-parameterized neural network, which outputs the mean $\muv$ and variance $\sigmav^2$ of the Gaussian distribution. 


After obtaining a user's latent representation $\zv$, we use the {\em generative} process to make predictions. In \cite{liang2018variational} a multinomial distribution is used to model the likelihood of items.
Specifically, to construct $p_{\thetav}(\xv | \zv)$, $\zv$ is transformed to produce a probability distribution $\piv$ over $M$ items, from which the interaction vector $\xv$ is assumed to
have been drawn:
\vspace{-0mm}
\begin{align} \label{eq_multi}
\xv \sim \mbox{Mult}  (\piv), ~~\mbox{with}~~\piv =\mbox{Softmax} (\exp\{ g_{\thetav} (\zv) \})
\end{align}
where $g_{\thetav}$ is a $\thetav$-parameterized neural network. 
%
The output $\piv$ is normalized via a softmax function to produce a probability vector $\piv \in \Delta^{M-1}$
(an ($M-1$)-simplex) over the entire item set.
%

\paragraph{Training Objective}~\hspace{-2mm}
Learning VAE parameters $\{\phiv, \thetav\}$ yields the following generalized objective:
%
%
\vspace{-1mm}
\begin{align} \label{eq_reg_elbo} \hspace{-2mm}
\Lcal_{\beta}(\xv; \thetav, \phiv) 
 \!=\! \Lcal_{E} \!+\!\beta \Lcal_{R}, 
 ~\text{with}~
\Lcal_{E}\!=\!-\E_{q_{\phiv}(\zv | \xv)} \big[  \log p_{\thetav}(\xv | \zv)   \big] 
 ~\text{and}~ \Lcal_{R}\!=\!\mbox{KL} (q_{\phiv}(\zv | \xv)    || p(\zv) ) 
\end{align}
where $\Lcal_{E}$ is the {\it negative log likelihood} (NLL) term, $\Lcal_{R}$ is the KL regularization term, and $\beta$ is a weighting hyper-parameter.
%
%
When $\beta=1$, we can lower-bound the log marginal likelihood of the data using \eqref{eq_reg_elbo} as
$
-\Lcal_{\beta=1}(\xv; \thetav, \phiv) \le \log p(\xv)
$.
This is commonly known as the {\it evidence lower bound} (ELBO) in variational inference~\cite{blei2017variational}. Thus \eqref{eq_reg_elbo} is the negative $\beta$-regularized ELBO. To improve the optimization efficiency, the {\it reparametrization trick}~\cite{kingma2013auto,rezende2014stochastic} is used to draw samples $\zv \sim  q_{\phiv}(\zv | \xv)$ to obtain an unbiased estimate of the ELBO, which is further optimized via stochastic optimization.
We call this procedure {\it maximum likelihood estimate (MLE)}-based training, as it effectively maximizes the (regularized) ELBO. The testing stage of VAEs for collaborative filtering is detailed in Section~\ref{sec:testing_vae} of Supplement.

\paragraph{Advantages of VAEs}
The VAE framework has a favorable characteristic that it is scalable to large datasets, by making use of amortized inference~\cite{gershman2014amortized}: the prediction for all users share the same procedure, which effectively requires evaluating two functions -- the encoder $f_{\phiv}(\cdot)$ and the decoder $g_{\thetav}(\cdot)$. This is much more efficient than most traditional latent factor collaborative filtering
models~\cite{paterek2007improving,hu2008collaborative,mnih2008probabilistic}, where a time-consuming optimization procedure is typically performed to obtain the latent
factor for a user who is not present in the training data. 
This makes the use of autoencoders particularly attractive in industrial applications, where fast prediction is important.
Interestingly, this amortized inference procedure reuses the same functions to answer related new problems. This is well aligned with collaborative filtering, where user preferences are analyzed by exploiting the similar patterns inferred from past experiences.
%
%

%

\begin{wrapfigure}{R}{0.52\textwidth}
\centering
	\vspace{-4mm}
	\begin{tabular}{c}		
		\includegraphics[width=7.00cm]{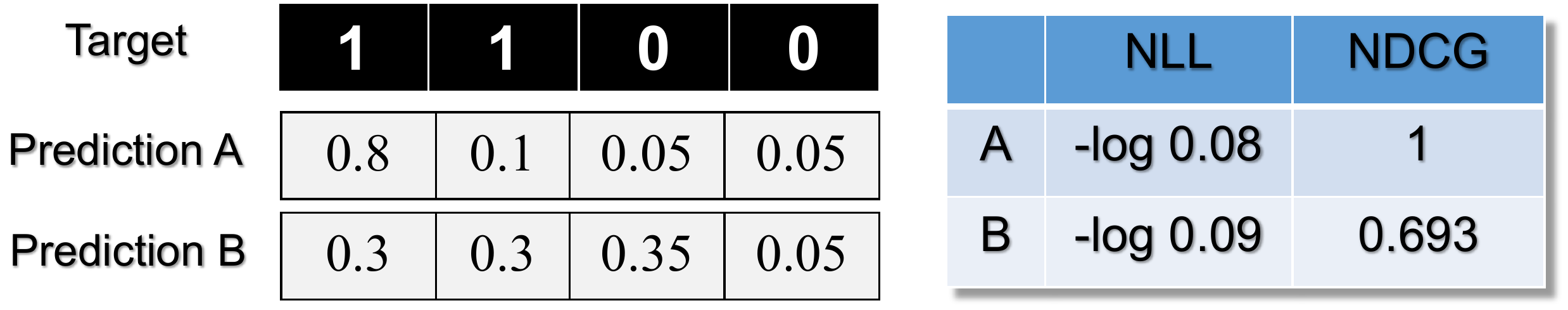} \\
	\end{tabular}
	\vspace{-0mm}
	\caption{\small Difference between MLE-based training loss and ranking-based evaluation. For A, $- 1 \! \times \!\log 0.8 - 1 \! \times \! \log 0.1 =- \log 0.08$; For B, $- 1 \! \times \! \log 0.4 - 1 \times \! \log 0.25 =- \log 0.10$. The multinomial NLL values disagree with the ground-truth that $A$ is a better recommendation than $B$, while NDCG values are coherent with the ground-truth.}
	\label{fig:example_div}
\vspace{-2mm}
\end{wrapfigure}

\paragraph{Pitfalls of VAEs} 
Among various likelihood forms, 
it was argued in~\cite{liang2018variational}  that multinomial likelihoods are a closer proxy to the ranking loss than the traditional Gaussian or logistic likelihoods.
Though simple and effective, the MLE procedure
may diverge with the ultimate goal in recommendation, of correctly suggesting the top-ranked items. Therefore, recommender systems are often evaluated using ranking-based measures, such as NDCG~\cite{jarvelin2002cumulated}.
To illustrate the divergence between MLE-based training and ranking-based evaluation, we provide an example in Figure~\ref{fig:example_div}. For the target $\xv=\{1,1,0,0\}$, two different predictions $A$ and $B$ are provided. In MLE, the training loss is the multinomial NLL: $-\xv \log \piv $, where $\piv$ is the predicted probability. From the NLL point of view, $B$ is a better prediction than $A$, because $B$ shows a lower loss than $A$. However, this apparently disagrees with our intuition that $A$ is better than $B$, because $A$ preserves the same ranking order with the target, while $B$ does not. Fortunately, the NDCG values correctly capture the true quality. This has inspired us to directly use ranking-based evaluation metrics to guide training.

\paragraph{From MLE to Ranking-Based Training }
The ranking loss is difficult to optimize, and previous work on its minimization has led practitioners to relaxations and approximations~\cite{weimer2008cofi}. Learning-to-ranking (L2R) methods have been studied in information retrieval~\cite{liu2009learning,li2014learning}, and some techniques can be extended to recommendation settings~\cite{rendle2009bpr,weston2013learning}.
Many L2R methods are essentially trained by optimizing a classification function, such as {\it Bayesian Personalized Ranking} (BPR)~\cite{rendle2009bpr} and {\it Weighted Approximate-Rank Pairwise} (WARP) model~\cite{weston2011wsabie}  (Detailed in Section~\ref{sec:two_l2r} of Supplement). 

When applying the traditional L2R methods to collaborative filtering, there are two potential issues: 
$(\RN{1})$ 
Each prediction is evaluated and optimized against the true ranking from scratch independently, and this process has to repeat for each new prediction, making L2R methods cumbersome for large-scale datasets.
$(\RN{2})$ The computation of the pairwise loss functions scales quadratically with the number of items, making many L2R methods inefficient to train on high-dimensional datasets.


\begin{figure*}[t!]
	\vspace{-0mm}\centering
	\begin{tabular}{c c}		
		\hspace{-0mm}
		\includegraphics[height=2.2cm]{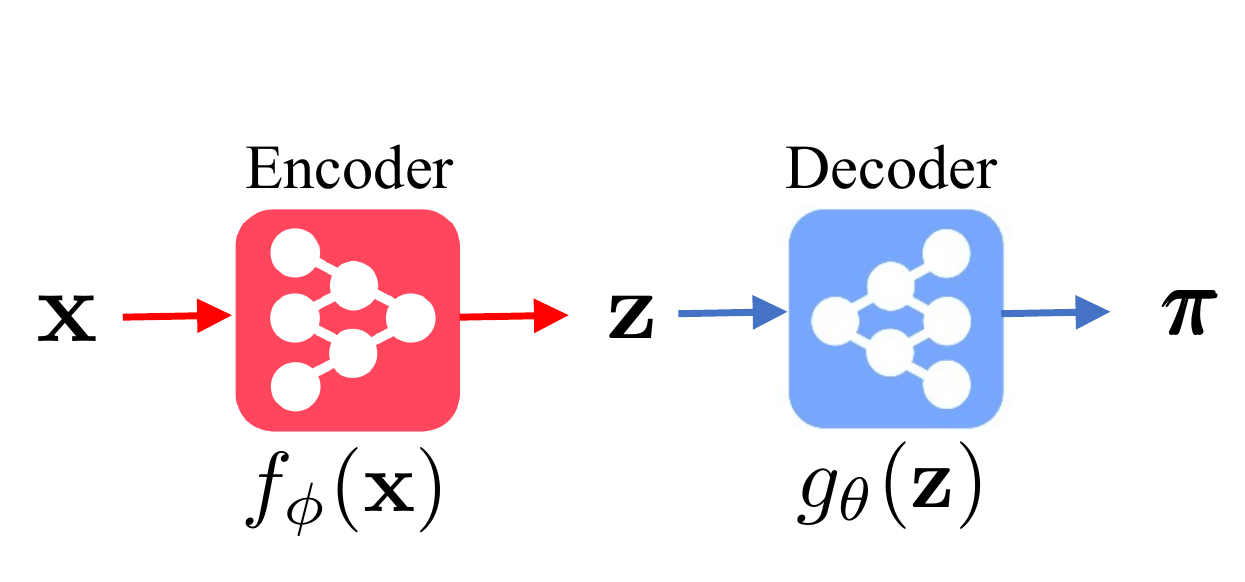} &
		\hspace{0mm}
		\includegraphics[height=2.2cm]{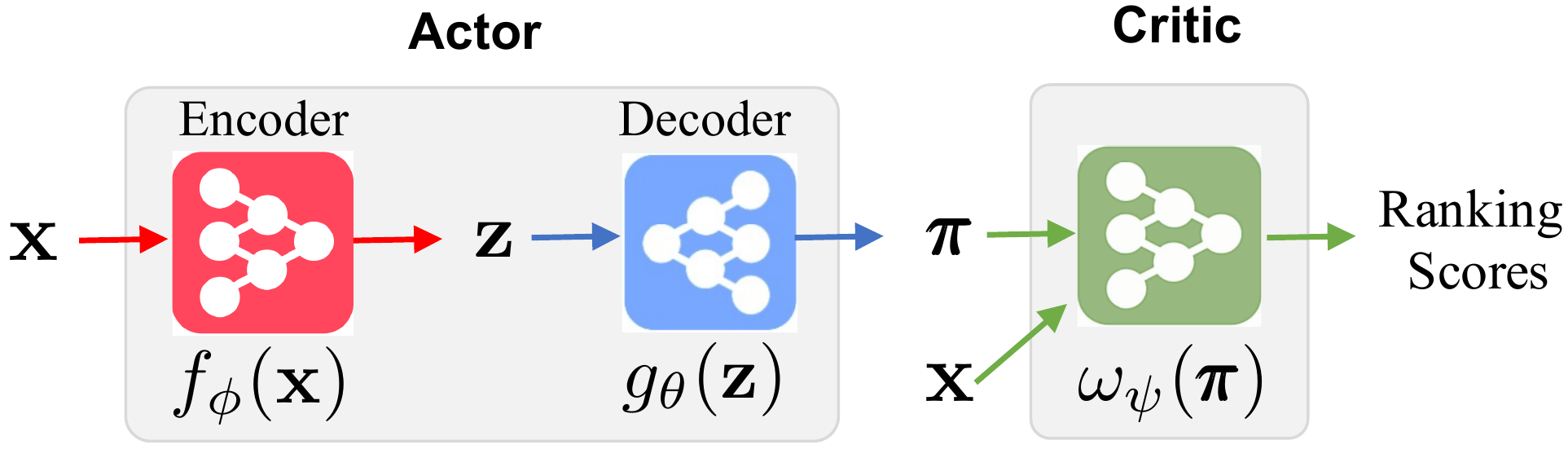} \\
		(a) Traditional auto-encoder paradigm \vspace{-0mm}   & 
		(b) Proposed actor-critic paradigm \hspace{-0mm} 
	\end{tabular}
	\vspace{-1mm}
	\caption{\small Illustration of learning parameters $\{\phiv,\thetav\}$ in the two different paradigms. (a) Learning with MLE, as in VAEs; (b) Learning with a learned ranking-critic. The {\it actor} can be viewed as the function composition of encoder $f_{\phiv}(\cdot)$ and $g_{\thetav}(\cdot)$ in VAEs. The {\it critic} mimics the ranking-based evaluation scores, so that it can provide ranking-sensitive feedback in the actor learning.}
	\vspace{-5mm}
	\label{fig:schemes}
\end{figure*}

\vspace{-3mm}
\section{Ranking-Critical Training}
\vspace{-3mm}
We introduce a novel actor-critic algorithm for ranking-based training, which we call {\it Ranking-Critical Training} (RaCT). 
The actor inherits the advantages of VAE to amortize the computation of collaborative filtering.
More importantly, the proposed neural-network-parameterized critic amortizes the computation of learning the ranking metrics, making RaCT scalable on large-scale datasets.



Any ranking-based evaluation metric can be considered as a ``black box'' function $\omega: \{ \piv; \xv, \bv \} \mapsto y \in [0, 1]$, which takes in the prediction $\piv$ to compare with the ground-truth $\xv$ (conditioned on the mask $\bv$), and outputs a scalar $y$ to rate the prediction quality.
Specifically, $\bv$ determines the items of interest in testing, \ie the items that are ``unobserved'' during inference.
As we are only interested in recovering the unobserved items in recommendation, we compute the ranking score of predicted items $\piv_p = \piv \odot (1-\bv)$ based on the ground-truth items $\xv_p = \xv \odot (1-\bv)$. 

One salient component of a ranking-based Oracle metric $\omega^*$ is to sort $\piv_p$. This operator is non-differentiable, rendering it impossible to directly use $\omega^*$ as the critic.
While REINFORCE~\cite{williams1992simple} may appear to be suited to tackle the non-differentiable problem, it suffers from  the issue of large estimate variance, as the collaborative filtering problem has a very large prediction space. 
This motivates consideration of a neural network to approximate the mapping executed by the Oracle.
This falls into the actor-critic paradigm in RL~\cite{sutton1998reinforcement}, and we borrow the idea for collaborative filtering. It consists of a policy network (actor) and value network (critic). The actor is trained to make a prediction (action) given the user's interaction history as the state. The critic predicts the value of each prediction, which we define as the task-specific reward, \ie the Oracle's output.
The value predicted by the critic is then used to train the actor. Under the assumption that the critic produces the exact values, the actor is trained based on an unbiased estimate of the gradient of the prediction value in terms of relevant ranking quality metrics. In Figure~\ref{fig:schemes}, we illustrate the actor-critic paradigm in (b), and the traditional auto-encoder shown in (a) can be used as the actor in our paradigm.

\paragraph{Naive critic}
Conventionally one may concatenate vectors $[\piv_p, \xv_p ]$ as input to a neural network, and train a network to output the measured ranking scores $y$.
However, this naive critic is impractical, and failed in our experiments. Our hypothesis is that since this network architecture has a huge number of parameters to train (as the input data layer is of length $2M$, where $M>10k$), it would require rich data for training. Unfortunately, this is impractical: $\{\piv, \xv\} \in \R^M$ are very high-dimensional, and hence it is too expensive to simulate enough data offline and then fit it to a scalar.

\vspace{-0mm}
\paragraph{Feature-based critic}
The naive critic hopes a deep network can discover structure from massive data by itself, leaving much valuable domain knowledge unused.
We propose a more efficient critic, by taking into account the structure underlined by the assumed likelihood in MLE~\cite{miyato2018cgans}. We describe our intuition and method below, and provide the justification from the perspective of adversarial learning in Section~\ref{sec:gan} of Supplement.


Consider the computation procedure of the evaluation metric as a function decomposition $ \omega  =    \omega_{0} \circ  \omega_{\psiv}$, including two steps:
%

\vspace{-2mm}
\begin{itemize} 
		\item
	    $ \omega_{0}:  \piv \mapsto  \hv $, feature engineering of prediction $ \piv $ into the {\it sufficient statistics} $\hv$ ; 
		\item 
		$ \omega_{\psiv}: \hv \mapsto \hat{y} $, neural approximation of the mapping from the statistics $\hv$ to the estimated ranking score $\hat{y}$, using a $\psiv$-parameterized neural network;
\end{itemize}

The success of this two-step critic largely depends on the effectiveness of the feature $\hv$. We hope feature $\hv$ is $(\RN{1})$  {\it compact} so that fewer parameters in the critic $ \omega_{\psiv} $ can simplify training; $(\RN{2})$ {\it easy-to-compute} so that training and testing is efficient; and $(\RN{3})$  {\it informative} so that the necessary information is preserved. 
We suggest to use a 3-dimensional vector as the feature, and leave more complicated feature engineering as future work. In summary, our feature is 
\begin{align} \label{eq_features}
\hv = [ \Lcal_{E} , |  \Hcal_0 |, |\Hcal_1 |], 
\end{align}
where
$(\RN{1})$  $\Lcal_{E}$ is the negative log-likelihood in~\eqref{eq_reg_elbo}, defined in the MLE training loss.
$(\RN{2})$  $| \Hcal_0 |$ is the number of unobserved items that a user will interact, with $\Hcal_0 = \{m| x_m = 1 ~\text{and}~b_m = 0\}$. 
$(\RN{3})$  $| \Hcal_1 |$ is the number of observed items that a user has interacted, with $\Hcal_1 = \{m| x_m = 1 ~\text{and}~b_m = 1\}$. 

The NLL characterizes the prediction quality of the actor's output $\pi$ against the ground-truth $\xv$ in an item-to-item comparison manner, \eg the inner product between two vectors $-\xv \log \piv $ as in the multinomial NLL~\cite{liang2018variational}. 
Note that the ideal optimum of the NLL yields a perfect match $\piv^{*} \propto \xv$, which also gives the perfect ranking scores. However, the {\it amortization gap} in amortized inference~\cite{kim2018semi,shu2018amortized} offsets the solutions obtained by VAEs from the ideal optimum. Fortunately, in recommendation we are only interested in ensuring that the order of top-ranking items are correct. This objective is easier to achieve, as it can be satisfied with some sub-optimal solutions of VAEs. Hence, we propose to adjust NLL to guide the actor towards them.

In practice, it is intractable to compute $\Lcal_{E}$, and a one-sample estimate is used for fast training
%
$
\Lcal_{E} \approx  -  \log p_{\thetav}(\xv | \zv), ~~\text{with}~~ 
\zv  \sim  q_{\phiv}(\zv | \xv)
$.
%
Note that $| \Hcal_0 |$ and $| \Hcal_1 |$ are user-specific, indicating the user's frequency to interact with the system, which can be viewed as side-information about the user. They are only used as features in training the critic to better approximate the ranking scores, and not in training the actor. Hence, we do not use additional information in the testing stage.

\paragraph{Critic Pre-training} 
Training a generic critic to approximate the ranking scores for all possible predictions is difficult and cumbersome. Furthermore, it is unnecessary. 
In practice, a critic only needs to estimate the ranking scores on the restricted domain of the current actor's outputs. Therefore, we train the critic offline on top of the pre-trained MLE-based actor.

To train the critic, we minimize the Mean Square Error (MSE) between the critic output and true ranking score $y$ from the Oracle:
\begin{align}~\label{eq_critic} 
\vspace{-3mm}
\Lcal_{C}(\hv, y; \psiv) = \| \omega_{\psiv} (\hv)  - y \|^2 ,
\end{align}
where the target $y$ is generated using its non-differential definition, which plays the role of ground truth simulator in training.

\paragraph{Actor-critic Training} 
Once the critic is well trained, we fix its parameters $\psiv$ and update the actor parameters $\{ \phiv, \thetav \}$ to maximize the estimated ranking score
\begin{align}~\label{eq_actor}  
\Lcal_{A}(\hv ; \phiv, \thetav) = \omega_{\psiv} (\hv),
\end{align}
where $\hv$ is defined in~\eqref{eq_features},
including NLL feature extracted from the prediction made in~\eqref{eq_reg_elbo}, together with count features. 
During back-propagation, the gradient of $\Lcal_{A}$ wrt the prediction $\piv$ is
$ 
\frac{\partial \Lcal_{A}}{\partial \piv } = 
\frac{\partial \Lcal_{A}}{\partial \hv }  \frac{\partial \hv}{\partial \piv }  .
$ 
It further updates the actor parameters, with the encoder gradient 
$ \frac{\partial \Lcal_{A}}{\partial \phiv } = 
\frac{\partial \Lcal_{A}}{\partial  \piv }  
\frac{\partial \piv}{\partial \phiv  } $
and the decoder gradient 
$ \frac{\partial \Lcal_{A}}{\partial \thetav } = 
\frac{\partial \Lcal_{A}}{\partial  \piv }  
\frac{\partial \piv}{\partial \thetav  } $.
Updating the actor changes its predictions, so we must update the critic to produce the correct ranking scores for its new input domain.

The full RaCT training procedure is summarized in Algorithm~1 in Supplement.
Stochastic optimization is used, where a batch of users 
 $\Ucal = \{\xv_i | i \in \Bcal \}$ is drawn at each iteration, with $\Bcal$ as a random subset of user index in $\{1, \cdots, N\}$. The pre-training of the actor in Stage 1 and the critic in Stage 2 are important; they provide good initialization to the actor-critic training in Stage 3 for fast convergence. Further, we provide an alternative interpretation to view our actor-critic approach in \eqref{eq_critic} and \eqref{eq_actor} from the perspective of adversarial learning~\cite{goodfellow2014generative} in Supplement. This can partially justify our choice of feature engineering.





\vspace{-3mm}
\section{Related Work}
\vspace{-3mm}
{\bf Deep Learning for Collaborative Filtering.}
To take advantage of the expressiveness of DNNs, there are many recent efforts focused on developing deep learning models for collaborative filtering~\cite{sedhain2015autorec,xue2017deep,he2018outer,he2018adversarial,zhang2017deep,chen2017attentive}. 
Early work on DNNs focused on explicit feedback settings~\cite{georgiev2013non,salakhutdinov2007restricted,zheng2016neural}, such as rating predictions. 
Recent research gradually recognized the importance of implicit feedback~\cite{wu2016collaborative,he2017neural,liang2018variational}, where the user's preference is not explicitly presented~\cite{hu2008collaborative}. This setting is more practical but challenging, and is the focus of our work.
Our method is closely related to three papers, on VAEs~\cite{liang2018variational}, collaborative denoising autoencoder (CDAE)~\cite{wu2016collaborative} and neural collaborative filtering (NCF)~\cite{he2017neural}.
%
%
%
CDAE and NCF may suffer from scalability issues: the model size grows linearly with both the number of users as well as items. 
The VAE~\cite{liang2018variational} alleviates this problem via amortized inference. 
Our work builds on top of the VAE, and improves it by optimizing to the ranking-based metric.

{\bf Learned Metrics in Vision \& Languages. } Recent research in computer vision and natural language processing has generated excellent results, by using learned metrics instead of hand-crafted metrics. Among the rich literature of generating realistic images via generative adversarial networks (GANs)~\cite{goodfellow2014generative,radford2015unsupervised,karras2017progressive,brock2018large}, our work is most similar to~ ~\cite{larsen2016autoencoding}, where the VAE objective~\cite{kingma2013auto,rezende2014stochastic} is augmented with the learned representations in the GAN discriminator~\cite{goodfellow2014generative} to better measure image similarities. 
For language generation, the discrepancy between word-level MLE training and sequence-level semantic evaluation has been alleviated with GANs or RL techniques~\cite{li2017adversarial,bahdanau2016actor,ren2017deep,yu2017seqgan,lin2017adversarial}. The RL approach directly optimizes the metric used at test time, 
and has shown improvement on various applications, including dialogue~\cite{li2016deep}, image captioning~\cite{rennie2017self} and translations~\cite{ranzato2015sequence}.
Despite the significant successes in vision and language analysis, there has been little if any research reported for directly learning the metrics with deep neural networks for collaborative filtering. Our work fills the gap, and we hope it inspires more research in this direction.

{\bf Learning to Rank (L2R).} The idea of L2R has existed for two decades in the information-retrieval community. The goal is to directly optimize against ranking-based evaluation metrics~\cite{liu2009learning,li2014learning}. Previous work on L2R employs objective relaxations~\cite{weimer2008cofi}. Some techniques can be extended to recommendation settings~\cite{rendle2009bpr,shi2010list,weston2013learning,shi2012tfmap,hidasi2018recurrent}. Many L2R methods in recommendation are essentially trained by optimizing a classification function, such as the popular pairwise L2R method BPR ~\cite{rendle2009bpr} and WARP~\cite{weston2011wsabie} described Section 2.1. One limitation is that they are computationally expensive when the number of items is large. To accelerate these approaches, cheap approximations are made in each training step, which results in degraded performance. In contrast, the proposed RaCT is efficient and scalable. In fact, the traditional L2R methods can be integrated into our actor-critic framework, yielding improved performance as shown in our experiments.

\vspace{-3mm}
\section{Experiments}
\vspace{-3mm}
\paragraph{Experimental Settings} 
We implemented our algorithm in TensorFlow.
The source code to reproduce the experimental results \& plots is included as Supplementary Material, and will be released on Github. 
We conduct experiments on three publicly available
large-scale datasets. These three ten-million-size datasets represent different item recommendation scenarios, including user-movie ratings and user-song play counts. This is the same set of user-item consumption datasets used in~\cite{liang2018variational}, and we keep the same pre-processing steps for fair comparison. 
The statistics of the datasets, evaluation protocols and hyper-parameters are summarized in Supplement. 
VAE~\cite{liang2018variational} is used as the baseline, which plays the role of our actor pre-training. The NCDG@100 ranking metric is used as the critic's target in training.

{\bf Baselines}~~~
We use ranking-critical training to improve the three MLE-based methods described in Section 2.1: VAE,
DAE, and MF. We also adapt a traditional L2R method as the actor in our framework.
The L2R loss is used to replace $\Lcal_{E}$ in \eqref{eq_features} to construct the feature.
Since WARP has been shown to perform generally better than BPR for collaborative filtering~\cite{kula2015metadata}, we only consider WARP in the experiments. 
%
We also compare our approaches with four representative state-of-the-art methods in collaborative filtering.
Two neural-network-based methods are CDAE~\cite{wu2016collaborative} and NCF~\cite{he2017neural}, and two linear models are Weighted MF~\cite{hu2008collaborative} and SLIM~\cite{ning2011slim}. 


\begin{figure*}[t!]
	\vspace{-0mm}\centering
	\begin{tabular}{c c c}		
		\hspace{-2mm}
		\includegraphics[height=3.4cm]{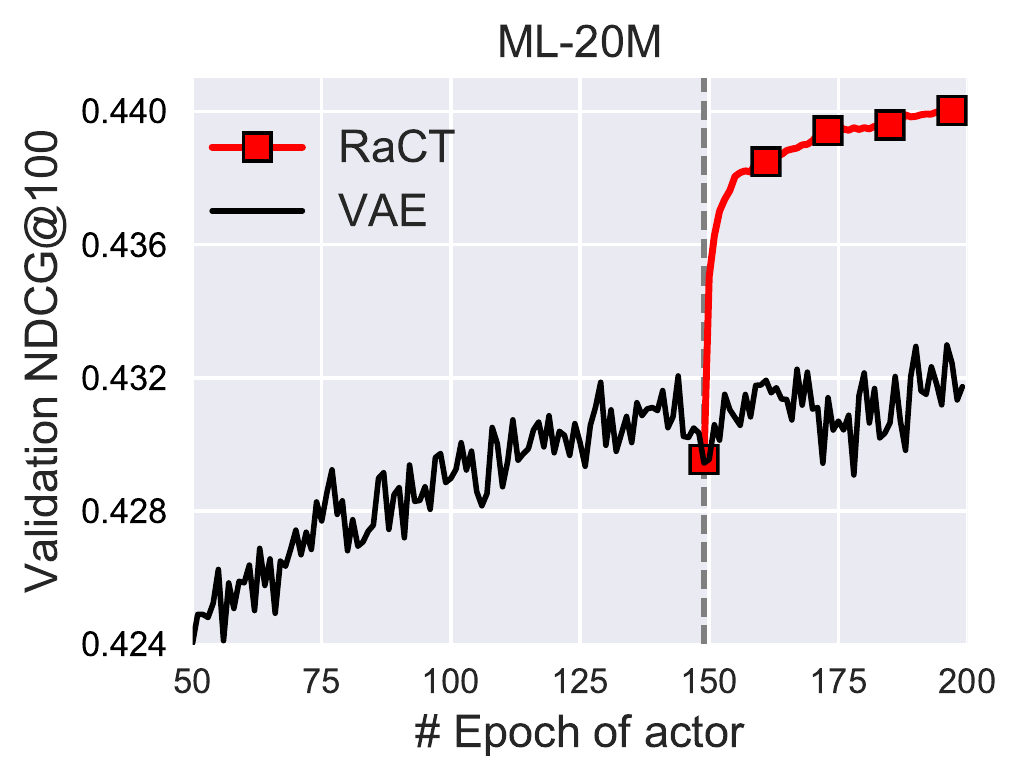} &
		\hspace{-4mm}
		\includegraphics[height=3.4cm]{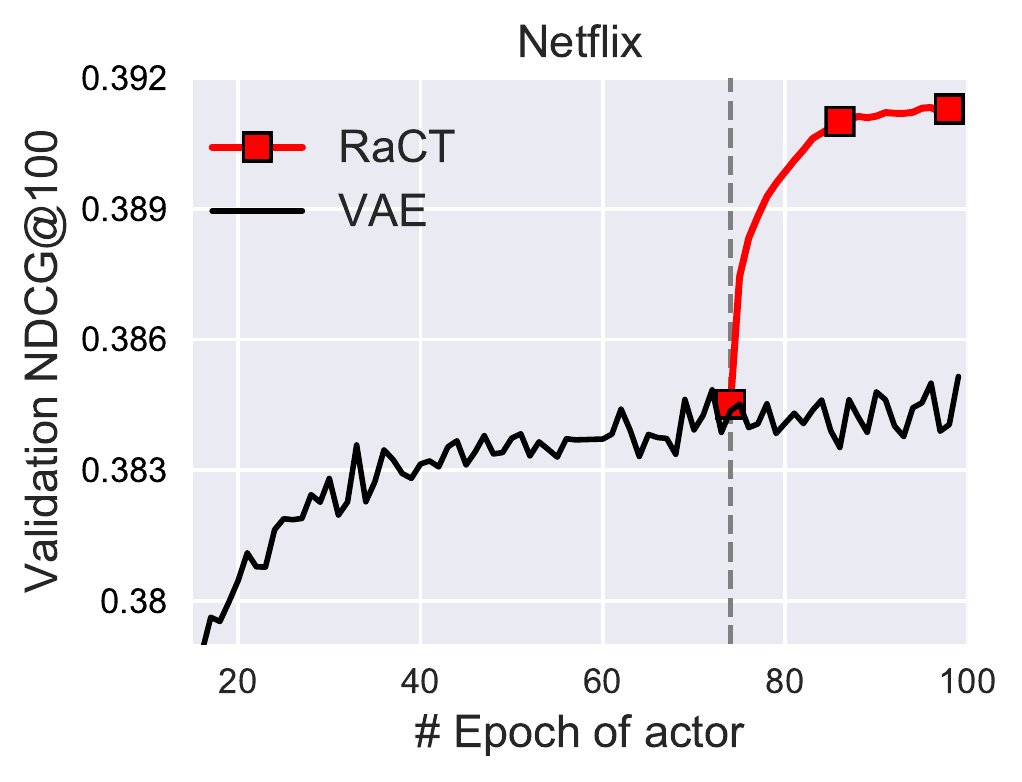} &
		\hspace{-4mm}
		\includegraphics[height=3.4cm]{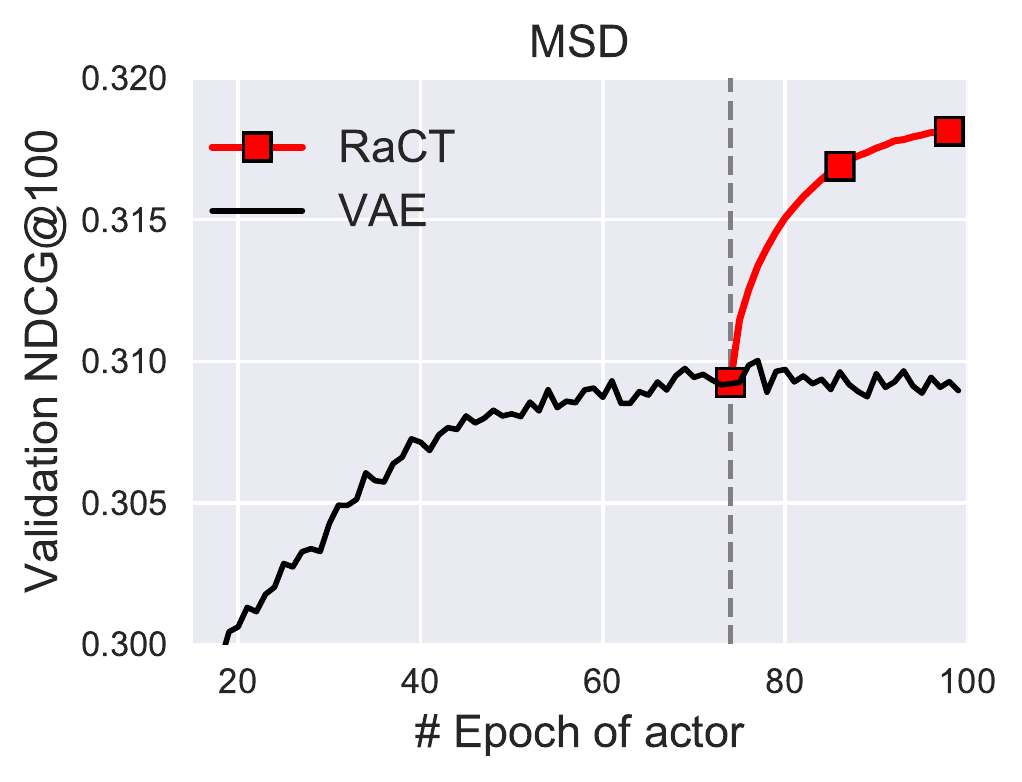}	
		\vspace{-2mm}
		\\
		(a) ML-20M dataset \vspace{-0mm}   & 
		(b) Netflix dataset \hspace{-0mm} &
		(c) MSD dataset\hspace{-0mm} \\ 
	\end{tabular}
	\vspace{-2mm}
	\caption{Performance improvement (NDCG@100) with RaCT over the VAE baseline.}
	\vspace{-4mm}
	\label{fig:improvement}
\end{figure*}

\subsection{Overall Performance of RaCT}
\vspace{-2mm}

\paragraph{Improvement over VAE} 
In Figure~\ref{fig:improvement}, we show the learning curves of RaCT and VAE on the validation set. The VAE converges to a plateau by the time that the RaCT finishes its actor pre-training stage, \eg 150 epochs on ML-20 dataset, after which the VAE's performance is not improving. In contrast, when the RaCT is plugged in, the performance shows a significant immediate boost. For the amount of improvement gain, RaCT takes only half the number of epochs that VAE takes in the end of actor pre-training. For example, RaCT takes 50 epochs (from 150 to 200) to achieve an improvement of 0.44-0.43 = 0.01, while VAE takes 100 epochs (from 50 to 150) to achieve an improvement of 0.43-0.424 = 0.006.


\begin{table*}[t!]
  \vspace{-0mm}
  \begin{adjustbox}{scale=.80,tabular=c|ccc|ccc|ccc}
    \toprule
    Dataset &
        \multicolumn{3}{ c|}{ML-20M}  & 
        \multicolumn{3}{ c|}{Netflix}  & 
        \multicolumn{3}{ c }{MSD}  \\ \hline
    Metric 
         & R@20 & R@50 & NDCG@100 
         & R@20 & R@50 & NDCG@100 
         & R@20 & R@50 & NDCG@100  \\
    \midrule       
    RaCT         
         & \textbf{0.403} & \textbf{0.543} & {\bf 0.434}
         & \textbf{0.357} & \textbf{0.450} & {\bf 0.392}
         & \textbf{0.268} & \textbf{0.364} & \textbf{0.319}  \\
    VAE$^{\ddag}$
         & 0.396 & 0.536 & 0.426 
         & 0.350 & 0.443 & 0.385 
         & 0.260 & 0.356 & 0.310  \\  
    WARP~\cite{weston2011wsabie}
        & 0.314 & 0.466	& 0.341
        & 0.270	& 0.365	& 0.306
        & 0.206	& 0.302	& 0.249 \\ 
    LambdaNet~\cite{burges2007learning}
        & 0.395	& 0.534	& 0.427
        & 0.352	& 0.441	& 0.386
        & 0.259	& 0.355	& 0.308  \\ 
         \hline
    VAE~\cite{liang2018variational}         
         & 0.395 & 0.537 & 0.426 
         & 0.351 & 0.444 & 0.386 
         & 0.266 & \textbf{0.364} & 0.316  \\ 
    CDAE~\cite{wu2016collaborative} 
         & 0.391 & 0.523 & 0.418
         & 0.343 & 0.428 & 0.376
         & 0.188 & 0.283 & 0.237 \\
    WMF~\cite{hu2008collaborative}         
         & 0.360 & 0.498 & 0.386
         & 0.316 & 0.404 & 0.351
         & 0.211 & 0.312 & 0.257 \\
    SLIM~\cite{ning2011slim} 
         & 0.370 & 0.495 & 0.401   
         & 0.347 & 0.428 & 0.379
         & --    & --    & --     \\
  \bottomrule
\end{adjustbox}
\vspace{-2mm}
  \caption{\small Comparison on three large datasets. The best testing set performance is reported. All numbers except RaCT and VAE$^{\ddag}$ are from~\cite{liang2018variational}, where VAE$^{\ddag}$ shows the VAE results based on our runs.}
  \label{tab:compare_sota}
\vspace{-4mm}
\end{table*}

{\bf Comparison with traditional L2R methods}
As examples of traditional L2R methods, we use WARP~\cite{weston2011wsabie} and LambdaNet~\cite{burges2007learning} as the ranking-critical objectives to optimize the VAE actor, to replace the last stage of RaCT. We observe that WARP and LambdaNet are roughly 2 and 10 times computationally expensive than RaCT per epoch, respectively. This is because the traditional L2R methods aim to minimize the number of incorrect pairs in ranking, which is not scalable in the high-dimensional datasets considered here.
More importantly, RaCT uses a neural network as the shared critic to amortize the computational cost among different predictions, while the traditional L2R methods do not have the amortized ranking-critical mechanism, and optimize each prediction independently. Table~\ref{tab:compare_sota} shows the results of RaCT, WARP and LambdaNet, using the same amount of wall-clock training time. We observe the trends that WARP degrades performance, and LambdaNet provides slight improvements if not worse. This is perhaps due to the poor approximation to the true ranking when the number of items is large.

{\bf Comparison with state-of-the-art}
In Table~\ref{tab:compare_sota}, we report our RaCT performance, and compare with state-of-the-art methods in terms of three evaluation metrics: NDCG@100, Recall@20, and Recall@50. 
We use the published code\footnote{\url{https://github.com/dawenl/vae_cf}} of~\cite{liang2018variational}, and reproduce the VAE as our actor pre-training. 
Our reproduced VAE results are very close to~\cite{liang2018variational} on the ML-20M and Netflix datasets, but slightly lower on the MSD dataset. The RaCT is built on top of our VAE runs, and consistently improves the baseline for all the evaluation metrics and datasets, as seen by comparing the rows RaCT and VAE$^{\ddag}$. 
The proposed RaCT also significantly outperforms other state-of-the-art methods, including VAE, CDAE, WMF and SLIM. 
Following~\cite{liang2018variational}, the comparison with NCF is shown on two small datasets due to its limited scalability, shown in Table~\ref{tab:compare_ncf} in Supplement. RaCT shows only slight improvements, perhaps because the estimate quality of the critic is poor when trained on small datasets.

\begin{wrapfigure}{R}{0.52\textwidth}
	\vspace{-7mm}\centering
	\begin{tabular}{c c}
		\hspace{-4mm}
		\includegraphics[height=3.5cm]{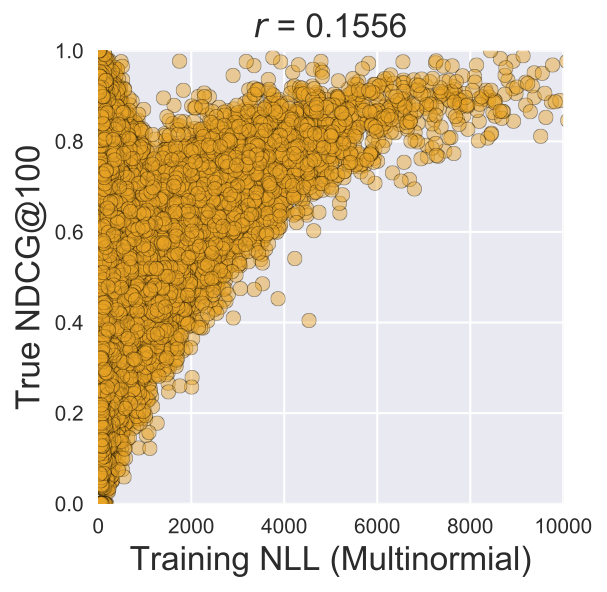} &
		\hspace{-7mm}
		\includegraphics[height=3.5cm]{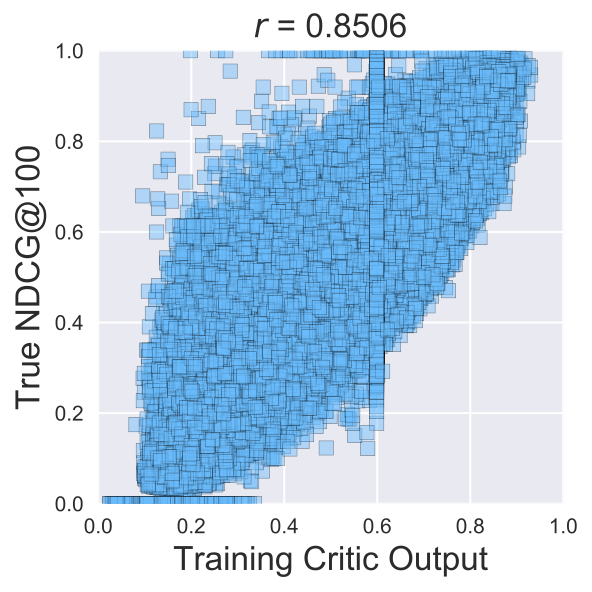}
		\\
		(a) MLE\vspace{-0mm}   & 
		(b) RaCT \hspace{-0mm} \\ 	
	\end{tabular}
	\vspace{-2mm}
	\caption{Correlation between the learning objectives (MLE or RaCT) and evaluation metrics on training.}
	\vspace{-2mm}
	\label{fig:correlation}
\end{wrapfigure}
\paragraph{Training/Evaluation Correlation} We visualize scatter plots between learning objectives and evaluation metric for all users on ML-20M dataset in Figure~\ref{fig:correlation}. The enlarged visualization is shown in Figure~\ref{fig:correlation_supp} of Supplement.
For training the objective, the VAE employs the NLL, while RaCT employs the learned NDCG metric. We ensure that the best model for each method is used: the model after actor pre-training (Stage 1) is used for NLL plots, and the model after the actor-critic alternative training (Stage 3) is used for RaCT plots.
The Pearson's correlation $r$ is computed. NLL exhibits low correlation with the target NDCG ($r$ is close to zero), while the learned metric in RaCT shows much higher positive correlation. It strongly indicates RaCT optimizes a more direct objective than an MLE approach. Further, NLL should in theory have a negative correlation with the target NDCG, as we wish that minimizing NLL can maximize NDCG. However, in practice it yields positive correlation. We hypothesize that this is because the number of interactions for each user may dominate the NLL values. In practice, the NLL value varies a lot; those with a higher number of interactions typically show both higher NLL and higher NDCG. That partially motivate us to consider the number of user interactions as features.

In Supplement, we study the generalization of RaCT trained with different ranking-metrics in Section~\ref{sec_metrics_supp}, and break down the performance improvement with different cut-off values of NDCG in Section~\ref{sec_cut_off_supp}, and with different number of interactions of $\Xmat$ in Section~\ref{sec_interactions_supp}.



\begin{table}[t!]
	\begin{minipage}{0.53\linewidth}
    \centering
      \begin{adjustbox}{scale=.93,tabular=l|ccc}
        \toprule
        Actor  & Before & After & Gain \\
        \midrule      
        VAE~\cite{liang2018variational}         
             & 0.4258 & 0.4339 & 8.09  \\ 
        VAE (Gaussian) 
             & 0.4202 & 0.4224 & 2.21  \\
        VAE  ($\beta = 0$)        
             & 0.4203 & 0.4255  & 5.17   \\
        VAE (Linear)
             & 0.4156  & 0.4162  &  0.53  \\ \hline
        DAE~\cite{liang2018variational}       
             & 0.4205 & 0.4214 & 0.87  \\
        MF~\cite{liang2018variational}     
             & 0.4159 & 0.4172 & 1.37  \\ \hline
        WARP~\cite{weston2011wsabie} 
             & 0.3123 & 0.3439  & 31.63   \\ 
      \bottomrule
    \end{adjustbox}
    \vspace{1mm}
      \caption{\small Performance gain ($\times 10^{-3}$) for various actors.}
      \vspace{-0mm}
      \label{tab:compare_actors}
	\end{minipage}\hfill
	\begin{minipage}{0.45\linewidth}
		\vspace{-2mm}
		\centering
    	\begin{tabular}{c}
    	    \hspace{-5mm}
    		\includegraphics[height=3.50cm]{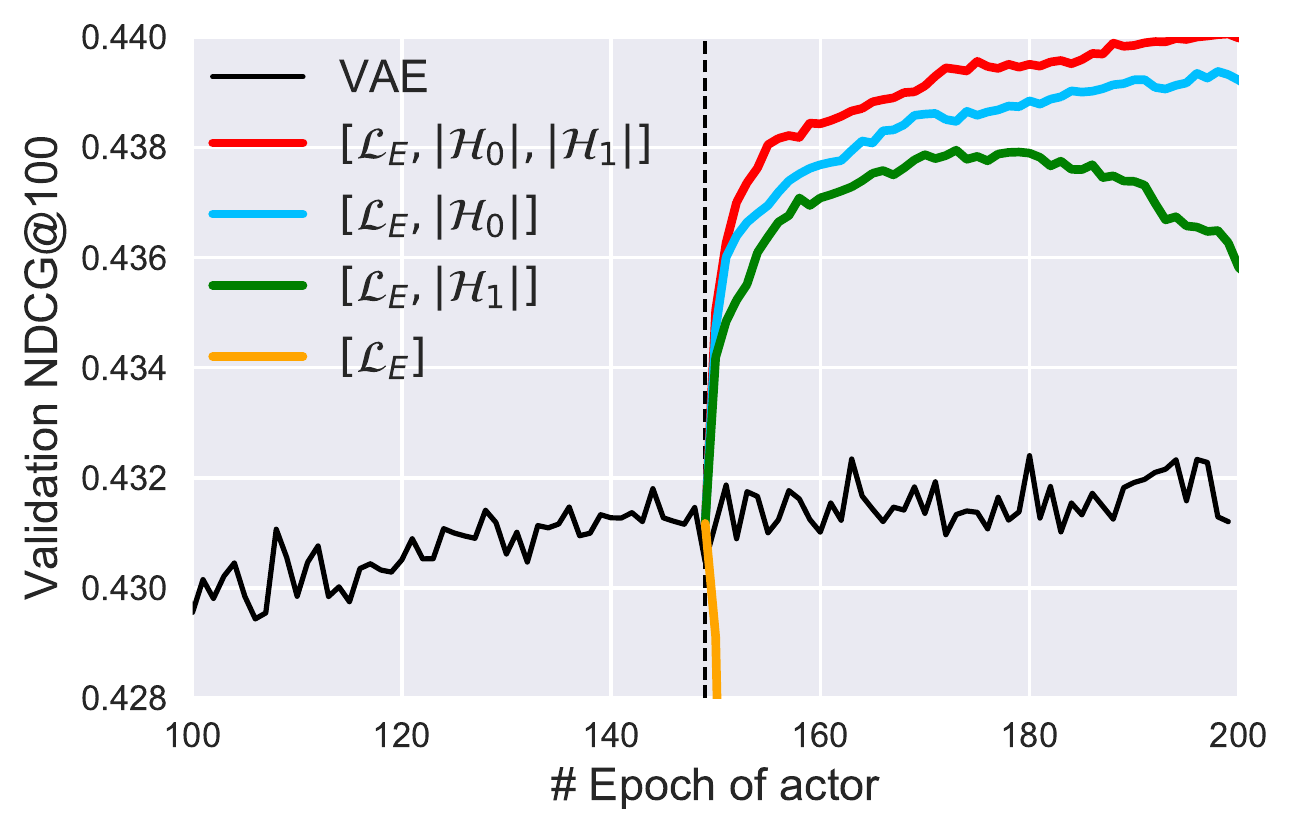} \\
    	\end{tabular}
    	\vspace{-1mm}
    	\captionof{figure}{\small Ablation study on features.
    	}
    	\vspace{-0mm}
    	\label{fig:feature_ablation}
	\end{minipage}
	\vspace{-4mm}
\end{table}
\subsection{What Actor Can Be Improved by RaCT?}
%
We investigate how RaCT performs with different actors. 
In RL, the policy plays a crucial role in the agent's performance. Similarly, we would like to study how different actor designs impact the RaCT performance.  The results are shown in Table~\ref{tab:compare_actors}. It shows the performance of before and after applying RaCT. The results on NDCG@100 are reported. The VAE, DAE and MF models follow the setups in~\cite{liang2018variational}.

We first modify one component of the VAE~\cite{liang2018variational} at a time, and check the change of performance improvement that RaCT can provide. 
(1) VAE (Gaussian): we change likelihood form from multinomial to Gaussian, and observe a smaller performance improvement. This shows the importance of having a closer proxy of ranking-based loss.
(2) VAE ($\beta=0$): we remove the KL regularization by setting $\beta=0$, and replace the posterior sampling with a delta distribution. We see a marginally smaller performance improvement. This compares a stochastic and deterministic policy.  The stochastic policy (\ie posterior sampling) provides higher exploration ability for the actor, allowing more diverse samples generated for the critic's training. This is essential for better critic learning. 
(3) VAE (Linear): we limit the expressive ability of the actor by using a linear encoder and decoder. This significantly degrades performance, and the RaCT cannot help much in this case. RaCT shows improvements for all MLE-based methods, including DAE and MF. It also shows significant improvement over WARP. 
Please see detailed discussion in Section \ref{sec:actors_supp} of Supplement.

\subsection{Ablation Study on Feature-based Critic} 
In Figure~\ref{fig:feature_ablation}, we investigate the importance of the features we designed in~\eqref{eq_features}. 
The full feature vector consists of three elements: 
$\hv = [ \Lcal_{E} , |  \Hcal_0 |, |\Hcal_1 |]$. 
$\Lcal_{E} $ is mandatory, because it links actor to the critic; removing it would break the back-propagation to train the actor. 
%
Results are gathered on the ML-20M dataset using the pre-trained VAE baseline. This ensures that the feature $\Lcal_{E}$ for the critic pre-training is always the same.
We carefully remove 
$|  \Hcal_0 |$ or $|\Hcal_1 |$ from $\hv$ at each time, and observe that it leads to performance degradation. In particular, removing $|  \Hcal_0 |$ results in a severe over-fitting issue. 
When both counts are removed, it shows an immediate performance drop, as depicted by the orange curve. Overall, the results indicate that all three features are necessary to our performance improvement.

\vspace{-2mm}
\section{Conclusion} 
\vspace{0mm}
We have proposed an actor-critic framework for collaborative filtering on implicit data. The critic learns to approximate the ranking scores, which in turn improves the traditional MLE-based nonlinear LVMs with the learned ranking-critical objectives.
To make it practical and efficient, we introduce a few techniques: a feature-based critic to reduce the number of learnable parameters, posterior sampling as exploration for better critic estimates, and pre-training of actor and critic for fast convergence.
The experimental results on three large-scale datasets demonstrate the superiority of the actor-critic approach, compared with state-of-the-art methods.


\medskip

\small

\bibliographystyle{unsrt}
\bibliography{neurips_2019}

\begin{thebibliography}{3}
\providecommand{\natexlab}[1]{#1}
\providecommand{\url}[1]{\texttt{#1}}
\expandafter\ifx\csname urlstyle\endcsname\relax
  \providecommand{\doi}[1]{doi: #1}\else
  \providecommand{\doi}{doi: \begingroup \urlstyle{rm}\Url}\fi

\bibitem[Bengio \& LeCun(2007)Bengio and LeCun]{Bengio+chapter2007}
Yoshua Bengio and Yann LeCun.
\newblock Scaling learning algorithms towards {AI}.
\newblock In \emph{Large Scale Kernel Machines}. MIT Press, 2007.

\bibitem[Goodfellow et~al.(2016)Goodfellow, Bengio, Courville, and
  Bengio]{goodfellow2016deep}
Ian Goodfellow, Yoshua Bengio, Aaron Courville, and Yoshua Bengio.
\newblock \emph{Deep learning}, volume~1.
\newblock MIT Press, 2016.

\bibitem[Hinton et~al.(2006)Hinton, Osindero, and Teh]{Hinton06}
Geoffrey~E. Hinton, Simon Osindero, and Yee~Whye Teh.
\newblock A fast learning algorithm for deep belief nets.
\newblock \emph{Neural Computation}, 18:\penalty0 1527--1554, 2006.

\end{thebibliography}


\begin{thebibliography}{65}
\providecommand{\natexlab}[1]{#1}
\providecommand{\url}[1]{\texttt{#1}}
\expandafter\ifx\csname urlstyle\endcsname\relax
  \providecommand{\doi}[1]{doi: #1}\else
  \providecommand{\doi}{doi: \begingroup \urlstyle{rm}\Url}\fi

\bibitem[Bahdanau et~al.(2016)Bahdanau, Brakel, Xu, Goyal, Lowe, Pineau,
  Courville, and Bengio]{bahdanau2016actor}
Dzmitry Bahdanau, Philemon Brakel, Kelvin Xu, Anirudh Goyal, Ryan Lowe, Joelle
  Pineau, Aaron Courville, and Yoshua Bengio.
\newblock An actor-critic algorithm for sequence prediction.
\newblock \emph{arXiv preprint arXiv:1607.07086}, 2016.

\bibitem[Bertin-Mahieux et~al.(2011)Bertin-Mahieux, Ellis, Whitman, and
  Lamere]{bertin2011million}
Thierry Bertin-Mahieux, Daniel~PW Ellis, Brian Whitman, and Paul Lamere.
\newblock The million song dataset.
\newblock In \emph{ISMIR}, 2011.

\bibitem[Bishop(2006)]{bishop2006pattern}
Christopher Bishop.
\newblock Pattern recognition and machine learning.
\newblock \emph{Pattern Recognition and Machine Learning}, 2006.

\bibitem[Blei et~al.(2017)Blei, Kucukelbir, and McAuliffe]{blei2017variational}
David~M Blei, Alp Kucukelbir, and Jon~D McAuliffe.
\newblock Variational inference: A review for statisticians.
\newblock \emph{Journal of the American Statistical Association}, 2017.

\bibitem[Brock et~al.(2018)Brock, Donahue, and Simonyan]{brock2018large}
Andrew Brock, Jeff Donahue, and Karen Simonyan.
\newblock Large scale gan training for high fidelity natural image synthesis.
\newblock \emph{arXiv preprint arXiv:1809.11096}, 2018.

\bibitem[Burges et~al.(2007)Burges, Ragno, and Le]{burges2007learning}
Christopher~J Burges, Robert Ragno, and Quoc~V Le.
\newblock Learning to rank with nonsmooth cost functions.
\newblock In \emph{NIPS}, 2007.

\bibitem[Cao et~al.(2007)Cao, Qin, Liu, Tsai, and Li]{cao2007listnet}
Zhe Cao, Tao Qin, Tie-Yan Liu, Ming-Feng Tsai, and Hang Li.
\newblock Learning to rank: From pairwise approach to listwise approach.
\newblock In \emph{International conference on machine learning}, 2007.

\bibitem[Chen et~al.(2017)Chen, Zhang, He, Nie, Liu, and
  Chua]{chen2017attentive}
Jingyuan Chen, Hanwang Zhang, Xiangnan He, Liqiang Nie, Wei Liu, and Tat-Seng
  Chua.
\newblock Attentive collaborative filtering: Multimedia recommendation with
  item-and component-level attention.
\newblock In \emph{Proceedings of the 40th International ACM SIGIR conference
  on Research and Development in Information Retrieval}, pp.\  335--344. ACM,
  2017.

\bibitem[Cremonesi et~al.(2010)Cremonesi, Koren, and
  Turrin]{cremonesi2010performance}
Paolo Cremonesi, Yehuda Koren, and Roberto Turrin.
\newblock Performance of recommender algorithms on top-{N} recommendation
  tasks.
\newblock In \emph{ACM Conference on Recommender systems}, 2010.

\bibitem[Dacrema et~al.(2019)Dacrema, Cremonesi, and Jannach]{dacrema2019we}
Maurizio~Ferrari Dacrema, Paolo Cremonesi, and Dietmar Jannach.
\newblock Are we really making much progress? a worrying analysis of recent
  neural recommendation approaches.
\newblock In \emph{ACM Conference on Recommender Systems}, 2019.

\bibitem[Finn et~al.(2016)Finn, Christiano, Abbeel, and
  Levine]{finn2016connection}
Chelsea Finn, Paul Christiano, Pieter Abbeel, and Sergey Levine.
\newblock A connection between generative adversarial networks, inverse
  reinforcement learning, and energy-based models.
\newblock \emph{arXiv preprint arXiv:1611.03852}, 2016.

\bibitem[Georgiev \& Nakov(2013)Georgiev and Nakov]{georgiev2013non}
Kostadin Georgiev and Preslav Nakov.
\newblock A non-iid framework for collaborative filtering with restricted
  boltzmann machines.
\newblock In \emph{International conference on machine learning}, pp.\
  1148--1156, 2013.

\bibitem[Gershman \& Goodman(2014)Gershman and Goodman]{gershman2014amortized}
Samuel Gershman and Noah Goodman.
\newblock Amortized inference in probabilistic reasoning.
\newblock In \emph{Proceedings of the Annual Meeting of the Cognitive Science
  Society}, 2014.

\bibitem[Goodfellow et~al.(2014)Goodfellow, Pouget-Abadie, Mirza, Xu,
  Warde-Farley, Ozair, Courville, and Bengio]{goodfellow2014generative}
Ian Goodfellow, Jean Pouget-Abadie, Mehdi Mirza, Bing Xu, David Warde-Farley,
  Sherjil Ozair, Aaron Courville, and Yoshua Bengio.
\newblock Generative adversarial nets.
\newblock In \emph{NIPS}, pp.\  2672--2680, 2014.

\bibitem[Goodfellow et~al.(2016)Goodfellow, Bengio, Courville, and
  Bengio]{goodfellow2016deep}
Ian Goodfellow, Yoshua Bengio, Aaron Courville, and Yoshua Bengio.
\newblock \emph{Deep learning}, volume~1.
\newblock MIT press Cambridge, 2016.

\bibitem[He et~al.(2017)He, Liao, Zhang, Nie, Hu, and Chua]{he2017neural}
Xiangnan He, Lizi Liao, Hanwang Zhang, Liqiang Nie, Xia Hu, and Tat-Seng Chua.
\newblock Neural collaborative filtering.
\newblock In \emph{Proceedings of the 26th International Conference on World
  Wide Web}, pp.\  173--182. International World Wide Web Conferences Steering
  Committee, 2017.

\bibitem[He et~al.(2018{\natexlab{a}})He, Du, Wang, Tian, Tang, and
  Chua]{he2018outer}
Xiangnan He, Xiaoyu Du, Xiang Wang, Feng Tian, Jinhui Tang, and Tat-Seng Chua.
\newblock Outer product-based neural collaborative filtering.
\newblock \emph{IJCAI}, 2018{\natexlab{a}}.

\bibitem[He et~al.(2018{\natexlab{b}})He, He, Du, and Chua]{he2018adversarial}
Xiangnan He, Zhankui He, Xiaoyu Du, and Tat-Seng Chua.
\newblock Adversarial personalized ranking for recommendation.
\newblock In \emph{The 41st International ACM SIGIR Conference on Research \&
  Development in Information Retrieval}, pp.\  355--364. ACM,
  2018{\natexlab{b}}.

\bibitem[Hu et~al.(2008)Hu, Koren, and Volinsky]{hu2008collaborative}
Yifan Hu, Yehuda Koren, and Chris Volinsky.
\newblock Collaborative filtering for implicit feedback datasets.
\newblock In \emph{Data Mining, 2008. ICDM'08. Eighth IEEE International
  Conference on}, pp.\  263--272. IEEE, 2008.

\bibitem[Ioffe \& Szegedy(2015)Ioffe and Szegedy]{ioffe2015batch}
Sergey Ioffe and Christian Szegedy.
\newblock Batch normalization: Accelerating deep network training by reducing
  internal covariate shift.
\newblock \emph{ICML}, 2015.

\bibitem[J{\"a}rvelin \& Kek{\"a}l{\"a}inen(2002)J{\"a}rvelin and
  Kek{\"a}l{\"a}inen]{jarvelin2002cumulated}
Kalervo J{\"a}rvelin and Jaana Kek{\"a}l{\"a}inen.
\newblock Cumulated gain-based evaluation of ir techniques.
\newblock \emph{ACM Transactions on Information Systems (TOIS)}, 2002.

\bibitem[Karras et~al.(2018)Karras, Aila, Laine, and
  Lehtinen]{karras2017progressive}
Tero Karras, Timo Aila, Samuli Laine, and Jaakko Lehtinen.
\newblock Progressive growing of gans for improved quality, stability, and
  variation.
\newblock \emph{ICLR}, 2018.

\bibitem[Kingma \& Ba(2014)Kingma and Ba]{kingma2014adam}
Diederik~P Kingma and Jimmy Ba.
\newblock Adam: A method for stochastic optimization.
\newblock \emph{ICLR}, 2014.

\bibitem[Kingma \& Welling(2013)Kingma and Welling]{kingma2013auto}
Diederik~P Kingma and Max Welling.
\newblock Auto-encoding variational bayes.
\newblock \emph{ICLR}, 2013.

\bibitem[Kula(2015)]{kula2015metadata}
Maciej Kula.
\newblock Metadata embeddings for user and item cold-start recommendations.
\newblock \emph{arXiv preprint arXiv:1507.08439}, 2015.

\bibitem[Larsen et~al.(2016)Larsen, S{\o}nderby, Larochelle, and
  Winther]{larsen2016autoencoding}
Anders Boesen~Lindbo Larsen, S{\o}ren~Kaae S{\o}nderby, Hugo Larochelle, and
  Ole Winther.
\newblock Autoencoding beyond pixels using a learned similarity metric.
\newblock In \emph{International Conference on Machine Learning}, pp.\
  1558--1566, 2016.

\bibitem[Li et~al.(2017)Li, Liu, Chen, Pu, Chen, Henao, and Carin]{li2017alice}
Chunyuan Li, Hao Liu, Changyou Chen, Yuchen Pu, Liqun Chen, Ricardo Henao, and
  Lawrence Carin.
\newblock Alice: Towards understanding adversarial learning for joint
  distribution matching.
\newblock In \emph{NIPS}, pp.\  5495--5503, 2017.

\bibitem[Li(2014)]{li2014learning}
Hang Li.
\newblock Learning to rank for information retrieval and natural language
  processing.
\newblock \emph{Synthesis Lectures on Human Language Technologies}, 7\penalty0
  (3):\penalty0 1--121, 2014.

\bibitem[Li et~al.(2016)Li, Monroe, Ritter, Galley, Gao, and
  Jurafsky]{li2016deep}
Jiwei Li, Will Monroe, Alan Ritter, Michel Galley, Jianfeng Gao, and Dan
  Jurafsky.
\newblock Deep reinforcement learning for dialogue generation.
\newblock \emph{arXiv preprint arXiv:1606.01541}, 2016.

\bibitem[Liang et~al.(2015)Liang, Zhan, and Ellis]{liang2015content}
Dawen Liang, Minshu Zhan, and Daniel~PW Ellis.
\newblock Content-aware collaborative music recommendation using pre-trained
  neural networks.
\newblock In \emph{ISMIR}, 2015.

\bibitem[Liang et~al.(2018)Liang, Krishnan, Hoffman, and
  Jebara]{liang2018variational}
Dawen Liang, Rahul~G Krishnan, Matthew~D Hoffman, and Tony Jebara.
\newblock Variational autoencoders for collaborative filtering.
\newblock \emph{WWW}, 2018.

\bibitem[Lin et~al.(2017)Lin, Li, He, Zhang, and Sun]{lin2017adversarial}
Kevin Lin, Dianqi Li, Xiaodong He, Zhengyou Zhang, and Ming-Ting Sun.
\newblock Adversarial ranking for language generation.
\newblock In \emph{NIPS}, pp.\  3155--3165, 2017.

\bibitem[Liu et~al.(2009)]{liu2009learning}
Tie-Yan Liu et~al.
\newblock Learning to rank for information retrieval.
\newblock \emph{Foundations and Trends{\textregistered} in Information
  Retrieval}, 3\penalty0 (3):\penalty0 225--331, 2009.

\bibitem[Mirza \& Osindero(2014)Mirza and Osindero]{mirza2014conditional}
Mehdi Mirza and Simon Osindero.
\newblock Conditional generative adversarial nets.
\newblock \emph{arXiv preprint arXiv:1411.1784}, 2014.

\bibitem[Miyato \& Koyama(2018)Miyato and Koyama]{miyato2018cgans}
Takeru Miyato and Masanori Koyama.
\newblock cgans with projection discriminator.
\newblock \emph{ICLR}, 2018.

\bibitem[Mnih \& Salakhutdinov(2008)Mnih and
  Salakhutdinov]{mnih2008probabilistic}
Andriy Mnih and Ruslan~R Salakhutdinov.
\newblock Probabilistic matrix factorization.
\newblock In \emph{NIPS}, 2008.

\bibitem[Nikolakopoulos \& Karypis(2019)Nikolakopoulos and
  Karypis]{nikolakopoulos2019recwalk}
Athanasios~N Nikolakopoulos and George Karypis.
\newblock Rec{W}alk: Nearly uncoupled random walks for top-{N} recommendation.
\newblock In \emph{ACM International Conference on Web Search and Data Mining},
  2019.

\bibitem[Nikolakopoulos et~al.(2019{\natexlab{a}})Nikolakopoulos, Berberidis,
  Karypis, and Giannakis]{nikolakopoulos2019personalized}
Athanasios~N Nikolakopoulos, Dimitris Berberidis, George Karypis, and
  Georgios~B Giannakis.
\newblock Personalized diffusions for top-{N} recommendation.
\newblock In \emph{ACM Conference on Recommender Systems}, 2019{\natexlab{a}}.

\bibitem[Nikolakopoulos et~al.(2019{\natexlab{b}})Nikolakopoulos, Kalantzis,
  Gallopoulos, and Garofalakis]{nikolakopoulos2019eigenrec}
Athanasios~N Nikolakopoulos, Vassilis Kalantzis, Efstratios Gallopoulos, and
  John~D Garofalakis.
\newblock Eigen{R}ec: generalizing pure{SVD} for effective and efficient
  top-{N} recommendations.
\newblock \emph{Knowledge and Information Systems}, 2019{\natexlab{b}}.

\bibitem[Ning \& Karypis(2011)Ning and Karypis]{ning2011slim}
Xia Ning and George Karypis.
\newblock Slim: Sparse linear methods for top-n recommender systems.
\newblock In \emph{2011 11th IEEE International Conference on Data Mining},
  pp.\  497--506. IEEE, 2011.

\bibitem[Paterek(2007)]{paterek2007improving}
Arkadiusz Paterek.
\newblock Improving regularized singular value decomposition for collaborative
  filtering.
\newblock In \emph{Proceedings of KDD cup and workshop}, volume 2007, pp.\
  5--8, 2007.

\bibitem[Pfau \& Vinyals(2016)Pfau and Vinyals]{pfau2016connecting}
David Pfau and Oriol Vinyals.
\newblock Connecting generative adversarial networks and actor-critic methods.
\newblock \emph{arXiv preprint arXiv:1610.01945}, 2016.

\bibitem[Radford et~al.(2015)Radford, Metz, and
  Chintala]{radford2015unsupervised}
Alec Radford, Luke Metz, and Soumith Chintala.
\newblock Unsupervised representation learning with deep convolutional
  generative adversarial networks.
\newblock \emph{arXiv preprint arXiv:1511.06434}, 2015.

\bibitem[Ranzato et~al.(2015)Ranzato, Chopra, Auli, and
  Zaremba]{ranzato2015sequence}
Marc'Aurelio Ranzato, Sumit Chopra, Michael Auli, and Wojciech Zaremba.
\newblock Sequence level training with recurrent neural networks.
\newblock \emph{ICLR}, 2015.

\bibitem[Ren et~al.(2017)Ren, Wang, Zhang, Lv, and Li]{ren2017deep}
Zhou Ren, Xiaoyu Wang, Ning Zhang, Xutao Lv, and Li-Jia Li.
\newblock Deep reinforcement learning-based image captioning with embedding
  reward.
\newblock \emph{CVPR}, 2017.

\bibitem[Rendle et~al.(2009)Rendle, Freudenthaler, Gantner, and
  Schmidt-Thieme]{rendle2009bpr}
Steffen Rendle, Christoph Freudenthaler, Zeno Gantner, and Lars Schmidt-Thieme.
\newblock Bpr: Bayesian personalized ranking from implicit feedback.
\newblock In \emph{Proceedings of the twenty-fifth conference on uncertainty in
  artificial intelligence}, pp.\  452--461. AUAI Press, 2009.

\bibitem[Rennie et~al.(2017)Rennie, Marcheret, Mroueh, Ross, and
  Goel]{rennie2017self}
Steven~J Rennie, Etienne Marcheret, Youssef Mroueh, Jarret Ross, and Vaibhava
  Goel.
\newblock Self-critical sequence training for image captioning.
\newblock In \emph{CVPR}, volume~1, pp.\ ~3, 2017.

\bibitem[Rezende et~al.(2014)Rezende, Mohamed, and
  Wierstra]{rezende2014stochastic}
Danilo~Jimenez Rezende, Shakir Mohamed, and Daan Wierstra.
\newblock Stochastic backpropagation and approximate inference in deep
  generative models.
\newblock \emph{ICML}, 2014.

\bibitem[Ricci et~al.(2015)Ricci, Rokach, and Shapira]{ricci2015recommender}
Francesco Ricci, Lior Rokach, and Bracha Shapira.
\newblock Recommender systems: introduction and challenges.
\newblock In \emph{Recommender systems handbook}, pp.\  1--34. Springer, 2015.

\bibitem[Salakhutdinov et~al.(2007)Salakhutdinov, Mnih, and
  Hinton]{salakhutdinov2007restricted}
Ruslan Salakhutdinov, Andriy Mnih, and Geoffrey Hinton.
\newblock Restricted boltzmann machines for collaborative filtering.
\newblock In \emph{Proceedings of the 24th international conference on Machine
  learning}, pp.\  791--798. ACM, 2007.

\bibitem[Sedhain et~al.(2015)Sedhain, Menon, Sanner, and
  Xie]{sedhain2015autorec}
Suvash Sedhain, Aditya~Krishna Menon, Scott Sanner, and Lexing Xie.
\newblock Autorec: Autoencoders meet collaborative filtering.
\newblock In \emph{Proceedings of the 24th International Conference on World
  Wide Web}, pp.\  111--112. ACM, 2015.

\bibitem[Sedhain et~al.(2016)Sedhain, Menon, Sanner, and
  Braziunas]{sedhain2016effectiveness}
Suvash Sedhain, Aditya~Krishna Menon, Scott Sanner, and Darius Braziunas.
\newblock On the effectiveness of linear models for one-class collaborative
  filtering.
\newblock In \emph{AAAI}, 2016.

\bibitem[Silver et~al.(2014)Silver, Lever, Heess, Degris, Wierstra, and
  Riedmiller]{silver2014deterministic}
David Silver, Guy Lever, Nicolas Heess, Thomas Degris, Daan Wierstra, and
  Martin Riedmiller.
\newblock Deterministic policy gradient algorithms.
\newblock 2014.

\bibitem[Steck(2019)]{steck2019ease}
Harald Steck.
\newblock Embarrassingly shallow autoencoders for sparse data.
\newblock 2019.

\bibitem[Su \& Khoshgoftaar(2009)Su and Khoshgoftaar]{su2009survey}
Xiaoyuan Su and Taghi~M Khoshgoftaar.
\newblock A survey of collaborative filtering techniques.
\newblock \emph{Advances in artificial intelligence}, 2009, 2009.

\bibitem[Sutton et~al.(1998)Sutton, Barto, Bach,
  et~al.]{sutton1998reinforcement}
Richard~S Sutton, Andrew~G Barto, Francis Bach, et~al.
\newblock \emph{Reinforcement learning: An introduction}.
\newblock MIT press, 1998.

\bibitem[Weimer et~al.(2008)Weimer, Karatzoglou, Le, and Smola]{weimer2008cofi}
Markus Weimer, Alexandros Karatzoglou, Quoc~V Le, and Alex~J Smola.
\newblock Cofi rank-maximum margin matrix factorization for collaborative
  ranking.
\newblock In \emph{NIPS}, pp.\  1593--1600, 2008.

\bibitem[Weston et~al.(2011)Weston, Bengio, and Usunier]{weston2011wsabie}
Jason Weston, Samy Bengio, and Nicolas Usunier.
\newblock Wsabie: Scaling up to large vocabulary image annotation.
\newblock In \emph{IJCAI}, volume~11, pp.\  2764--2770, 2011.

\bibitem[Weston et~al.(2013)Weston, Yee, and Weiss]{weston2013learning}
Jason Weston, Hector Yee, and Ron~J Weiss.
\newblock Learning to rank recommendations with the k-order statistic loss.
\newblock In \emph{Proceedings of the 7th ACM conference on Recommender
  systems}, pp.\  245--248. ACM, 2013.

\bibitem[Williams(1992)]{williams1992simple}
Ronald~J Williams.
\newblock Simple statistical gradient-following algorithms for connectionist
  reinforcement learning.
\newblock \emph{Machine learning}, 8\penalty0 (3-4):\penalty0 229--256, 1992.

\bibitem[Wu et~al.(2016)Wu, DuBois, Zheng, and Ester]{wu2016collaborative}
Yao Wu, Christopher DuBois, Alice~X Zheng, and Martin Ester.
\newblock Collaborative denoising auto-encoders for top-n recommender systems.
\newblock In \emph{Proceedings of the Ninth ACM International Conference on Web
  Search and Data Mining}, 2016.

\bibitem[Xia et~al.(2008)Xia, Liu, Wang, Zhang, and Li]{xia2008listmle}
Fen Xia, Tie-Yan Liu, Jue Wang, Wensheng Zhang, and Hang Li.
\newblock Listwise approach to learning to rank - theory and algorithm.
\newblock 2008.

\bibitem[Xue et~al.(2017)Xue, Dai, Zhang, Huang, and Chen]{xue2017deep}
Hong-Jian Xue, Xinyu Dai, Jianbing Zhang, Shujian Huang, and Jiajun Chen.
\newblock Deep matrix factorization models for recommender systems.
\newblock In \emph{IJCAI}, pp.\  3203--3209, 2017.

\bibitem[Zhang et~al.(2017)Zhang, Yao, and Sun]{zhang2017deep}
Shuai Zhang, Lina Yao, and Aixin Sun.
\newblock Deep learning based recommender system: A survey and new
  perspectives.
\newblock \emph{arXiv preprint arXiv:1707.07435}, 2017.

\bibitem[Zheng et~al.(2016)Zheng, Tang, Ding, and Zhou]{zheng2016neural}
Yin Zheng, Bangsheng Tang, Wenkui Ding, and Hanning Zhou.
\newblock A neural autoregressive approach to collaborative filtering.
\newblock \emph{arXiv preprint arXiv:1605.09477}, 2016.

\end{thebibliography}

\normalsize

\appendix
\newpage
\title{\Large Appendix: \\
RaCT}

\section{Testing stage of VAEs for Collaborative Filtering}~\label{sec:testing_vae}
We focus on studying the performance of various models under strong generalization~\cite{liang2015content} as in~\cite{liang2018variational}. All users are split into training/validation/test sets. The models are learned using the entire interaction history of the users in the training set. To evaluate, we use a part of the interaction history from held-out (validation
and test) users to infer the user-level representations from the model, and compute quality metrics by quantifying how well the
model ranks the rest of the unseen interaction history from the held-out users.
Specifically, for a held-out user with the full history $\xv$, we take $\xv_h= \xv \odot \bv$ offline using the randomly generated mask $\bv$.  $\xv_h$ is then frozen as the testing input, and is fed into various trained models during the evaluation stage to get the prediction $\hat{\piv}$. The recovered  interaction $\bar{\xv}= \hat{\piv} \odot ( 1 - \xv_h)$ for the masked seen part is then evaluated by ranking-based metrics.

\section{Background on Traditional Learning-to-Rank Methods}~\label{sec:two_l2r}
Formally, the {\it Bayesian Personalized Ranking} (BPR)~\cite{rendle2009bpr} loss for the $n$-th user is
\begin{align}~\label{eq:objective_bpr}
\Lcal_{\mbox{\scriptsize BPR} } = 
\sum_{i \in \Kcal_{+} }  \sum_{j \in \Kcal_{-} } 
 \sigma ( \pi_{nj}  - \pi_{ni} ),
\end{align}
where $\sigma(\cdot)$ is the sigmoid function, $\Kcal_{+}$ denotes the set of items that the user has interacted with before, and $\Kcal_{-}$ denotes the complement item set.

The {\it Weighted Approximate-Rank Pairwise} (WARP) model~\cite{weston2011wsabie} has been shown to perform better than BPR for implicit feedback~\cite{kula2015metadata}:
\begin{align} \label{eq:objective_warp}
\Lcal_{ \mbox{\scriptsize WARP} } = 
\sum_{ni \in \Kcal_{+} }  \sum_{j \in \Kcal_{-} }
 w(r_i)
\mbox{max} (0, 1  + \pi_{nj} - \pi_{ni}  )  ),
\end{align}
where $w(\cdot)$ is a weighting function for different ranks, and $r_i$ is the rank for the $i$-th item for $n$-th user. A common choice of weighting function $w(\cdot)$ for optimizing NDCG is
$ w(r) = \sum_{i=1}^{r} \alpha_i, ~~\text{with}~~ 
\alpha_i = 1/i$. WARP improves BPR by the weights $w(\cdot)$ and the margin between positive and negative items.

\section{Pseudo-code for RaCT}~\label{sec:code}
We summarize the full training procedure of RaCT in Algorithm 1. 
\begin{algorithm}[h!]
	\SetKwInOut{Input}{Input}
	\caption{Our full ranking-critical training with stochastic optimization.}
	\Input{
		{\small Interaction matrix $\Xmat$; 
		Actor parameters (encoder $\phiv$ and decoder $\thetav$), Critic parameters $\psiv$}. }
	{\bf Initialize}: Randomly initialize weights  $\phiv$, $\thetav$ and $\psiv$ \\
	{\small \color{blue}  \tcc{Stage~1: Pretrain~~the~~actor~~via~~MLE} }
	 \While{not converged do}{
	 Sample a batch of users $\Ucal$; \\
	 Update $\{\thetav, \phiv\}$ with gradient
	 $\frac{\partial \Lcal_{\beta}}{\partial \thetav } $
     and 
     $\frac{\partial \Lcal_{\beta}}{\partial\phiv } $
	 in~\eqref{eq_reg_elbo};
	 } 

	\vspace{2mm}
	{\color{blue} \small  \tcc{Stage~2: Pretrain~the~critic~via~MSE} }	

	 \While{not converged do}{
	 Sample a batch of users $\Ucal$; \\
	 Construct features $\hv$ in~\eqref{eq_features} and target $y$ from the Oracle;\\ 
	 Update $\psiv$ with gradient 
	 $\frac{\partial \Lcal_{C}}{\partial\psiv } $
	 in~\eqref{eq_critic};
	 }

	\vspace{2mm}
	{\small \color{blue} \tcc{Stage 3: Alternative~training~of~actor~and~critic}} 	
	\For {$t = 1, 2, \ldots, T $} {
	    Sample a batch of users $\Ucal$; \\
		{\small \color{blue} \tcc{~Actor~~step} } 
	 Update $\{\thetav, \phiv\}$ with gradient
	 $\frac{\partial \Lcal_{A}}{\partial \thetav } $
     and 
     $\frac{\partial \Lcal_{A}}{\partial\phiv } $
	 in~\eqref{eq_actor};
		
		{\small \color{blue}  \tcc{~Critic~~step} } 
     Construct features $\hv$ in~\eqref{eq_features} and target $y$ from the Oracle;\\ 
	 Update $\psiv$ with gradient 
	 $\frac{\partial \Lcal_{C}}{\partial\psiv } $
	 in~\eqref{eq_critic};

	}
	\label{alg:actor_critic}
\end{algorithm}


\section{Interpretation with GANs}~\label{sec:gan}
We can view our actor-critic approach in \eqref{eq_critic} and \eqref{eq_actor} from the perspective of Generative Adversarial Networks (GANs). GANs constitute a framework to construct a {\it generator} $G$ that can mimic a target distribution, and have achieved significant success in generating realistic images~\cite{goodfellow2014generative,radford2015unsupervised,karras2017progressive,brock2018large}. The most distinctive feature
of GANs is the {\it discriminator} $D$ that evaluates the divergence between the current generator distribution and the target distribution~\cite{goodfellow2014generative,li2017alice}. The GAN learning procedure performs iterative training between the discriminator and generator, with the discriminator acting as an increasingly meticulous critic to refine the generator. In our work, the actor can be interpreted as the generator, while the critic can be viewed as the discriminator. 

Note that GANs and actor-critic models learn the metric functions~\cite{finn2016connection}, and it has been shown in~\cite{pfau2016connecting} that GANs can be viewed as actor-critic in an environment where the actor cannot affect the reward. This is exactly our setup. 
One key difference is that we know the Oracle metric, and the critic is trained to mimic the Oracle's behaviour. 

Conditioned on interaction history $\xv_h$ corrupted from $\xv$, the actor predicts the distribution parameter $\piv$ over items, which further constructs the likelihood $p(\xv | \piv)$. We use $q$ to designate the data empirical distribution, the target conditional is $q( \xv | \piv )$. It can be formulated as the standard adversarial loss for the conditional GAN~\cite{mirza2014conditional}. 
It has been shown that the optimal critic~\cite{goodfellow2014generative,li2017alice} for a conditional GAN can be represented as the log likelihood ratio
\begin{align}\label{eq:log_ratio}
R^*(\piv, \xv) = \log 
\frac{ q(\xv | \piv ) }{ p(\xv | \piv ) }
\end{align}
In the collaborative filtering setup, we often make the assumptions that $p(\xv | \piv )$ are simple distributions, such as multinomial in VAEs~\cite{liang2018variational} and Gaussian in MF. This simplification allows the parameterization of critic following the following form~\cite{miyato2018cgans}:
\begin{align}\label{eq:log_ratio_ip}
R^*(\piv, \xv) = 
\xv^{\top} \Vmat \nu  (\piv) + \Cmat
\end{align}
where $\xv$ is the target, $\nu(\piv)$ is a layer of the critic with input $\piv$, and $\Vmat$ and $\Cmat$ are the parameters to learn.  Most notably, this formulation introduces the prediction information via an inner product, as opposed to concatenation. The form~\eqref{eq:log_ratio_ip} is indeed the form we proposed for NLL feature $\xv \log \piv$, with $\Vmat = \Imat$ and $\nu (\cdot) = \log (\cdot)$. $\Cmat$ includes the normalizer for the prediction probability~\cite{miyato2018cgans}, which is related to the count features in~\eqref{eq_features}.

\section{Experimental setup} 

\subsection{Datasets}
We conduct experiments on three publicly available
datasets. Table~\ref{tab:dataset} summarizes the statistics of the data. These three ten-million-size datasets represent different item recommendation scenarios, including user-movie ratings and user-song play counts. This is the same set of medium- to large-scale user-item consumption datasets used in~\cite{liang2018variational}, and we keep the same pre-processing steps for fair comparison. 

\begin{enumerate}
\item {\bf MovieLens-20M (ML-20M)}: This is the user-movie rating data collected from a movie recommendation service\footnote{\url{https://grouplens.org/datasets/movielens/20m/}}. The data is binarized by keeping ratings of four or higher and setting other entries as unobserved. Only users who have watched at least
five movies are considered.
\item {\bf Netflix Prize (Netflix)}: This is the user-movie rating data from the Netflix Prize\footnote{\url{https://www.netflixprize.com/}}. Similarly to ML-20M, the data is binarized by keeping ratings of four or higher, and only users who have
watched at least five movies are kept.
\item {\bf Million Song Dataset (MSD)}: This is the user-song play count data from the Million Song Dataset~\cite{bertin2011million}. We
binarize play counts, and keep users who have listened to at least 20 songs as well as songs that are listened to by at least 200 users.
\end{enumerate}

\subsection{Evaluation Protocol}
In the testing stage, we get the predicted ranking by sorting the multinomial probability $\piv_p$. For each user, we compare the predicted ranking of the held-out items with their true ranking. 
Two ranking-based metrics are considered, Recall@R and the truncated NDCG (NDCG@R), where $R$ is the cut-off hyper-parameter.
While Recall@R considers all items ranked within the first $R$ to be equally important, NDCG@R uses a monotonically increasing discount to emphasize the importance of higher ranks versus lower ones. 

Formally, we define $m(r)$ as the item at rank $r$, and $\Hcal_0 $ as the held-out unobserved items that a user will interact. 
%
\begin{align}~\label{eq_dcg} \vspace{-3mm}
{\rm DCG@R} = \sum_{r=1}^{R}  \frac{ 2^{\delta[ m(r) \in \Hcal_0 ] } - 1 }{\log (r+1) } .
\end{align}
By dividing DCG@R by its best possible value, we obtain NDCG@R in $[0, 1]$.
\begin{align}~\label{eq_recall}\vspace{-3mm}
{\rm Recall@R} = \sum_{r=1}^{R}  \frac{  \delta[ m(r) \in \Hcal_0 ] }{\min(R, |\Hcal_0|) } .
\end{align}
The denominator normalizes Recall@R in $[0, 1]$, with maximum value $1$ corresponding to the case that all relevant items are ranked in the top $R$ positions.
%

\begin{table}[t!]
  \caption{Summary Statistics of datasets after all pre-processing steps. {\it Interactions$\#$} is the number of non-zero entries.  {\it Sparsity$\%$} refers to the percentage of zero entries in the user-item interaction matrix $X$. {\it Items$\#$} is the number of total items. {\it HO$\#$} is the number of validation/test users held out of the total number of users in the 5th column {\it Users$\#$}.}
  \label{tab:dataset}
  \centering
  \begin{tabular}{c|c|c|cc|c}
    \toprule
    Dataset & 
    Interaction\#  & \hspace{-0mm}Sparsity\% &  Item\#  &  User\#  
    &  \hspace{-1mm}  HO\# \\
    \midrule       
    ML-20M &  10.0M & 99.64\% & 20,108 & 136,677 & 10K \\
   Netflix &  56.9M & 99.31\% & 17,769 & 463,435 & 40K \\
    MSD    &  33.6M & 99.86\% & 41,140 & 571,355 & 50K \\
  \bottomrule
\end{tabular}
\end{table}
\begin{table}[t!]
\caption{Network architectures. The arrow indicates the flow between two layers. For each layer, we show the number of units on top of its following activation function. $\mathtt{BN}$ indicates Batch Normalization.}
\label{tab:network}
\centering
\begin{tabular}{c|c|c}
\toprule
\multicolumn{2}{ c|}{Networks}  & Architectures   \\ \hline
\multirow{2}{*}{Actor} 
& Encoder & 
$
\begin{matrix}
M \\ \mathtt{Linear}
\end{matrix} 
\rightarrow  
\begin{matrix}
600 \\ \mathtt{Tanh}
\end{matrix} 
\rightarrow 
\begin{matrix}
200 \\ \mathtt{Linear \&~Exp}
\end{matrix} 
$   \\ \cline{2-3}
& Decoder & $
\begin{matrix}
200 \\ \mathtt{Linear}
\end{matrix}  
\rightarrow  
\begin{matrix}
600 \\ \mathtt{Tanh}
\end{matrix} 
\rightarrow 
\begin{matrix}
M \\ \mathtt{Softmax}
\end{matrix} 
$   \\ \hline
\multicolumn{2}{ c|}{Critic}  & 
$
\begin{matrix}
3 \\  \mathtt{BN}
\end{matrix}
\rightarrow  
\begin{matrix}
100 \\  \mathtt{ReLU}
\end{matrix}
\rightarrow 
\begin{matrix}
100 \\  \mathtt{ReLU}
\end{matrix}
\rightarrow  
\begin{matrix}
10 \\  \mathtt{ReLU}
\end{matrix}
\rightarrow  
\begin{matrix}
1 \\  \mathtt{Sigmoid}
\end{matrix}
$
\\ 
\bottomrule
\end{tabular}
\end{table}

\subsection{Experiment Hyper-parameters}
We set hyper-parameters by following~\cite{liang2018variational} for comparisons. For VAE, the dimension of the latent representation is 200. 
When KL regularization is removed ($\beta=0$), \ie for DAE and MF, we instead apply $\ell_2$ regularization ($0.01$) on weights to prevent overfitting.
Adam optimizer~\cite{kingma2014adam} is used, with batch size of $|\Bcal|=500$ users. 
For ML-20M, the actor is pre-trained for 150 epochs, and alternative training for 50 epochs. On the other two datasets, the actor is pre-trained for 75 epochs, and alternative training for 25 epochs. The critic is pre-trained for 50 epochs for all three datasets. The alternative training has equal update frequency for actor and critic.
This schedule ensures that we the have the same total number of actor training epochs as ~\cite{liang2018variational}: 200 epochs for ML-20M, 100 epochs for the other two datasets.

A fully-connected (FC) architecture is used for all networks, as detailed in Table~\ref{tab:network}. Please refer to~\cite{goodfellow2016deep} for the activation functions. Batch Normalization~\cite{ioffe2015batch} is used to normalize the input features, 
because the magnitude of the inputs (NLL) change as training progresses. 
The encoder outputs the mean and variance of the varational distribution; the variance is implemented via an exponential function.

\begin{table}[t!]
  \caption{Summary of training schedule hyper-parameters. $\beta_{\max}$ indicates the maximum value of $\beta$. In the actor pre-training stage, the number of epochs used for increasing and fixing $\beta$ are shown in row 3 and 4, respectively. }
    \centering
  \label{tab:beta}
  \begin{tabular}{c|c|c|c}
    \toprule
    Dataset & ML-20M & Netflix & MSD \\ 
    \midrule  
    $\beta_{\max}$  &  0.2 &  0.2  &  0.1 \\
    \# epochs for annealing & 100 & 75 & 75 \\
     \# epochs for fixing & 50 & 0  & 0\\ \midrule  
    \# epochs for actor pre-training & 150 & 75 & 75 \\
    \# epochs for critic pre-training & 50 & 50 & 50 \\
    \# epochs for alternative training & 50 & 25 & 25 \\
  \bottomrule
\end{tabular}
\end{table}

\section{Additional Experimental Results}

\begin{figure}[t!]
	\vspace{-0mm}\centering
	\begin{tabular}{c c}		
		\hspace{-4mm}
		\includegraphics[height=5.0cm]{figs/correlation_objective/correlation_scatter_training_softmax.png} &
		\hspace{-5mm}
		\includegraphics[height=5.0cm]{figs/correlation_objective/correlation_scatter_training_ract.png} \\
		(a) Training NLL\vspace{-0mm}   & 
		(b) Training NDCG \hspace{-0mm}  \\ 		
		\hspace{-5mm}
		\includegraphics[height=5.0cm]{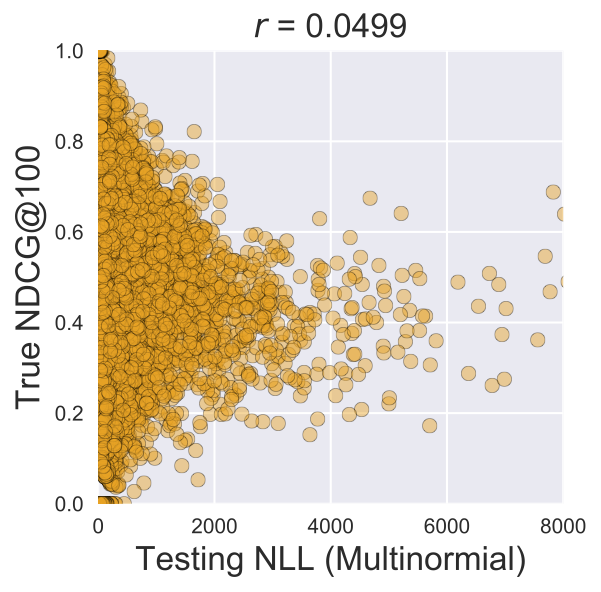} &
		\hspace{-5mm}
		\includegraphics[height=5.0cm]{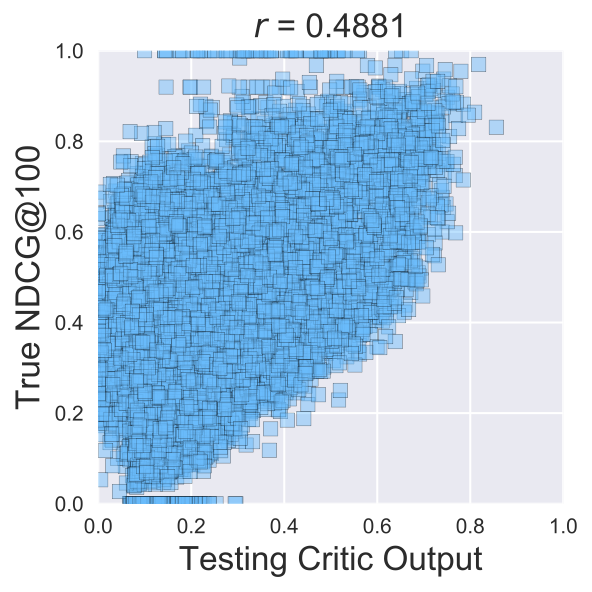}			
		\\
		(c) Testing NLL \vspace{-0mm}   & 
		(d) Testing NDCG\hspace{-0mm} \\ 	
	\end{tabular}
	\vspace{-2mm}
	\caption{Correlation between the learning objectives (NLL or RaCT) and evaluation metrics NDCG.}
	\vspace{-2mm}
	\label{fig:correlation_supp}
\end{figure}

\subsection{Generalization across ranking metrics} ~\label{sec_metrics_supp}
To study the generalization ability of RaCT, we consider training the critic against Recall@100, in addition to NDCG@100. The only difference is that Recall treats each item as equally important, while NDCG treats the higher ranking items as more important. The results are shown in Table~\ref{tab:generalize_metrics}. Indeed, the RaCT gets slightly better testing Recall values when trained against the Recall metric, and the reverse holds for NDCG. 
More importantly, RaCT allows generalization across different ranking metrics: all testing metric values are significantly improved when trained against either Recall or NDCG.

\begin{table}[t!]
  \caption{Performance trained with different metrics. (ML-20M)}
  \label{tab:generalize_metrics}
    \centering
  \begin{tabular}{c|ccc}
    \toprule
      Training   & \multicolumn{3}{c}{Testing}   \\ \cline{2-4}
         & Recall@20 & Recall@50 & NDCG@100   \\
    \midrule   
       RaCT (Recall@100)  & \textbf{0.40316} & \textbf{0.54317} & 0.43392 \\
       RaCT (NDCG@100)  & 0.40269 & 0.54304 & \textbf{0.43395} \\
       VAE   & 0.39623 & 0.53632 & 0.42586 \\
  \bottomrule
\end{tabular}
\end{table}

Following~\cite{liang2018variational}, we compare with NCF on two small datasets, ML-1M (6,040 users, 3,704 items) and Pinterest (55,187 users, 9,916 items). This is because the prediction stage of NCF is slow, due to a lack of amortized inference as in VAE. We use their publicly available datasets and metrics for fair comparison.
The results are evaluated with a small cut-off value $R$, to only study the highly ranked items: NDCG@10 and Recall@10. The performance are compared in Table~\ref{tab:compare_ncf}. 
Our observation that DAE performs better than VAE on these two datasets is consistent with~\cite{liang2018variational}. 
In general, RaCT shows higher improvement when a larger dataset (Pinterest), or a stochastic actor (VAE) is considered. This is because the sizes of the two datasets are relatively small, the critic can be better trained when more samples are observed.
On the larger Pinterest dataset, the auto-encoder variants perform better than NCF by a big margin, and our RaCT further boosts the performance.

\begin{table}[t!]
  \caption{Comparison between our RaCT with NCF on two small datasets. NCF results are from~\cite{liang2018variational}.}
  \vspace{0mm}
  \label{tab:compare_ncf}
    \centering
  \begin{tabular}{c|c|c|cc|cc}
    \toprule
   Dataset & Metric  &  NCF & DAE  & RaCT & VAE & RaCT \\ 
    \midrule   
    \multirow{2}{*}{ML-1M}
     & Recall@10   & 0.705 & 0.722 & 0.722 & 0.704 & 0.706 \\
     & NDCG@10     & 0.426 & 0.446 & 0.446 & 0.433 & 0.434 \\ \hline
    \multirow{2}{*}{Pinterest}
     & Recall@10    & 0.872 & 0.886 & 0.887 & 0.873 & 0.878 \\
     & NDCG@10      & 0.551 & 0.580 & 0.581 & 0.564 & 0.568  \\
  \bottomrule
\end{tabular}
\vspace{-0mm}
\end{table}

\begin{figure}[t!] \centering
	\vspace{-0mm}
	\begin{tabular}{c}
	    \hspace{-0mm}
		\includegraphics[height=4.00cm]{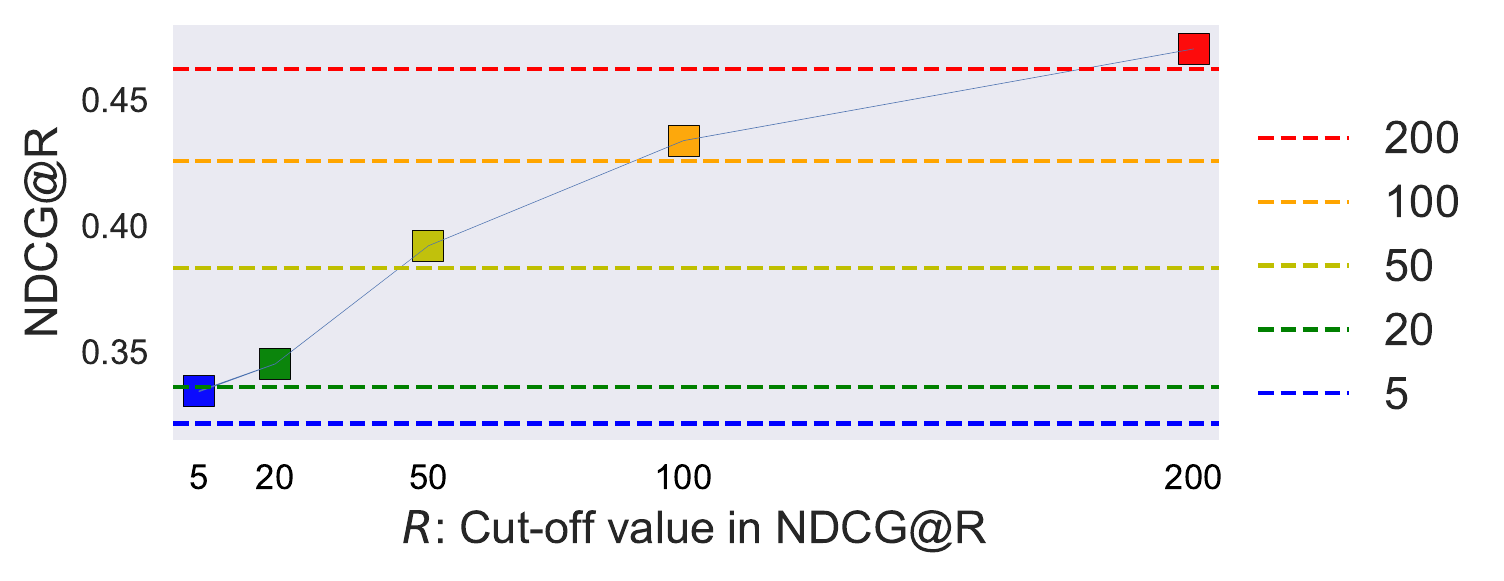} \\
	\end{tabular}
	\vspace{-0mm}
	\caption{The improvement at various cut-off value R in evaluation. Given a specific R, the dashed line shows the VAE, and square dot shows the RaCT.}
	\vspace{-0mm}
	\label{fig:cut_off}
\end{figure}

\subsection{Breakdown analysis for different cut-off values}~\label{sec_cut_off_supp} 
NDCG@100 only reflects the ranking quality at the cut-off value $R=100$. \ie the top-100 ranking items. To study the ranking quality at different range of the predicted list, we consider a large range of $R$, and report the corresponding NDCG values. We consider $R= 5, 20, 50, 100, 200$, and report the results in Figure~\ref{fig:cut_off}. The NDCG@R values are improved for various R, though the critic is trained against NDCG@100. This is because the NDCG metrics of different R are highly correlated, the RaCT can generalize across them.

\subsection{Breakdown analysis for different number of interactions}~\label{sec_interactions_supp} 
\begin{figure}[t!]
	\vspace{-2mm}\centering
	\begin{tabular}{c c }		
		\hspace{-4mm}
		\includegraphics[height=3.2cm]{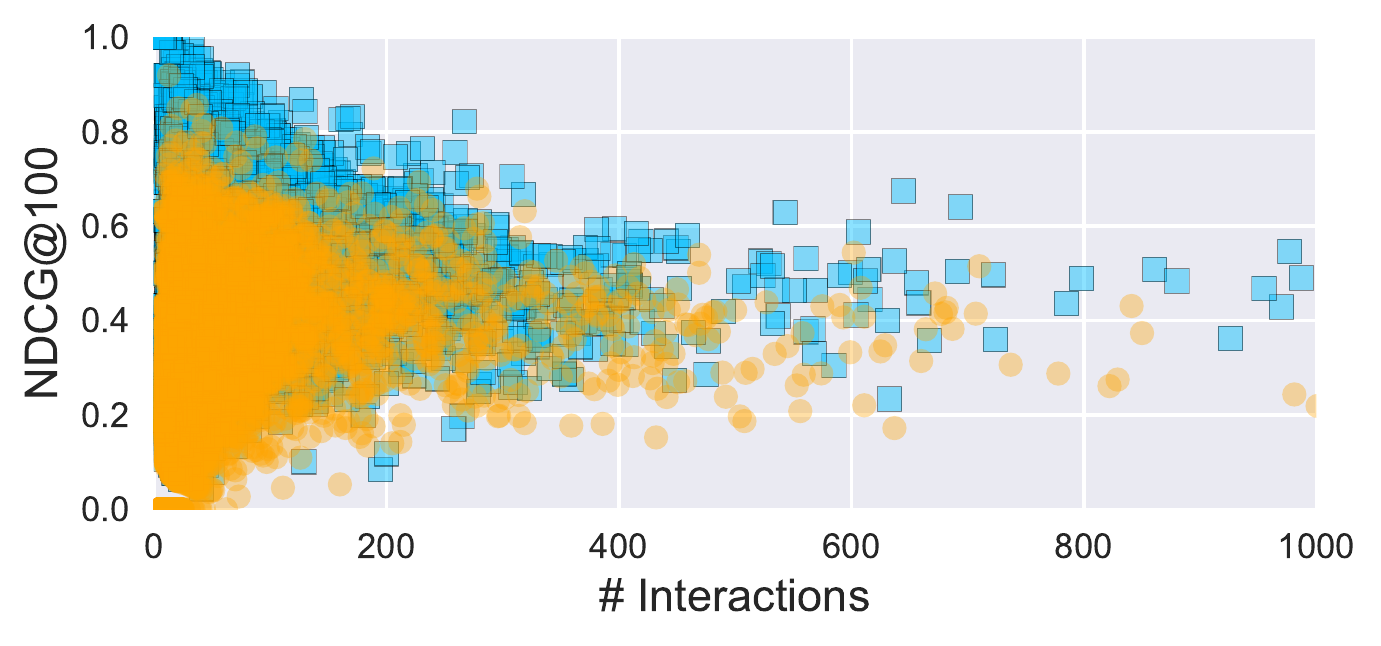} \vspace{-0mm} & 
		\hspace{-4mm}
		\includegraphics[height=3.1cm]{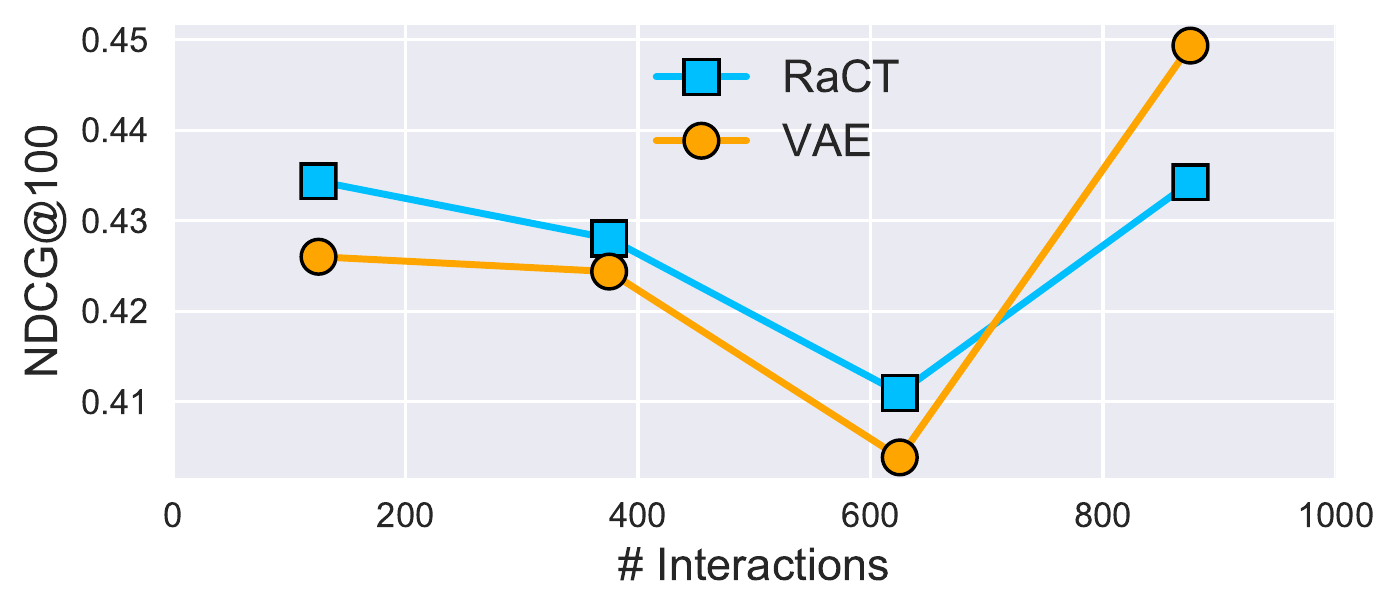} \vspace{-0mm} \\
		(a) Scatter plot \vspace{0mm}   &
		(b) NDCG mean 
	\end{tabular}
	\vspace{-3mm}
	\caption{Improvement breakdown over different user interactions. (a) Scatter plot between NDCG@100 and activity levels. Note only $\#$ interactions $\le$ 1000 is visualized, there is a long tail ($>$1000) in the distribution. (b) Comparison of the mean NDCG@100 values for four user groups.  }
	\vspace{-2mm}
	\label{fig:breakdown}
\end{figure}

In Figure~\ref{fig:breakdown}, we show performance improvement across increasing user interactions.
We use ML-20M dataset for this case study.
The \# interactions is the number of items
each user interacts with (ground-truth), indicating the user's activity level. Figure~\ref{fig:breakdown}(a) shows the scatter plots between NDCG@100 values and various number of interactions on the testing dataset, for both VAE and our RaCT methods. RaCT generally improves VAE for a large range of user interactions. We further categorize the users in four groups according to their number of interactions: $<\!\!250$, $250\!-\!500$, $501\!-\!750$, $>\!\!750$, and plot the mean of NDCG@100 values for two methods in Figure~\ref{fig:breakdown}(b). RaCT improves VAE except for users with high activity level ($>\!\!750$). This is probably because the number of the most active users is small, as observed in Figure~\ref{fig:breakdown}(a). It yields a lack of training data for critic learning, which potentially hurts the performance.

\subsection{On the performance improvement of actors via RaCT.}\label{sec:actors_supp}
We also consider the two other auto-encoder variants used in~\cite{liang2018variational} as the actor. (1) The DAE in~\cite{liang2018variational} chooses a smaller architecture $M \rightarrow 600 \rightarrow M$, which achieves better performance than the larger architecture as in our VAE ($\beta=0$) by prevent over-fitting. While we observe the same result, it is interesting to note that the VAE ($\beta=0$) shows a much larger improvement gain than DAE~\cite{liang2018variational} when trained with our RaCT technique, and eventually significantly outperforms the latter. This shows that the additional modeling capacity is necessary to capture the more complex relationship in prediction, when the goal is ranking rather than MLE. (2) The MF in~\cite{liang2018variational} employs a Gaussian likelihood, which also gets slight improvement with the RaCT. Overall, we can conclude that the RaCT method improves all the MLE-based variants.

We also use ranking-loss-based WARP as the actor. 
For the large datasets considered in this paper, calculating the full WARP-loss for each user is impractically slow. We derive a simple approximation to WARP which runs in quasilinear time to the number of items. 
Even so, it takes around 30 minutes per epoch on ML-20M dataset, roughly 30 times slower than the VAE. WARP yields the score 0.312, which is lower than other baseline methods. This is consistent with the studies in~\cite{liang2018variational,sedhain2016effectiveness}. However, when RaCT is applied, WARP gets a significant improvement; in fact, the largest improvement gain of all the actors. This indicates the RaCT is a more direct and effective approach for learning to rank on large datasets.

\end{document}